\documentclass[10pt,journal,compsoc]{IEEEtran}

\ifCLASSOPTIONcompsoc
  \usepackage[nocompress]{cite}
\else
  \usepackage{cite}
\fi

\ifCLASSINFOpdf
\else
\fi
\usepackage{rotating} 
\usepackage{enumitem}
\usepackage{array}
\usepackage{multirow}
\usepackage{subfig}
\usepackage{scrextend}
\usepackage{xspace}
\usepackage{url}
\usepackage{amsfonts}
\usepackage{bm}
\usepackage{xcolor}
\usepackage{amsmath}
\usepackage{textcomp}
\usepackage{makecell}
\usepackage{graphicx}  
\usepackage{hyperref}
\usepackage{amsthm}
\usepackage{color}
\usepackage{xcolor}
\usepackage{colortbl}
\newcommand{\mname}{\texttt{AICare}\xspace}

\usepackage{CJKutf8}

\theoremstyle{definition}
\newtheorem{definition}{Definition}
\newtheorem{problem}{Problem}

\begin{document}

\title{\huge Mortality Prediction with Adaptive Feature Importance Recalibration for Peritoneal Dialysis Patients: 

\Large a deep-learning-based study on a real-world longitudinal follow-up dataset}

\author{
Liantao~Ma$^{\#}$, Chaohe~Zhang$^{\#}$, Junyi~Gao$^{\#}$, Xianfeng~Jiao, Zhihao~Yu, Xinyu Ma, Yasha~Wang*, Wen~Tang*, Xinju~Zhao, Wenjie~Ruan, and~Tao Wang
\IEEEcompsocitemizethanks{
\IEEEcompsocthanksitem L. Ma, C. Zhang, X. Jiao, Z. Yu, X. Ma, Y. Wang, were with Peking University, CN.\protect\\
L. Ma, C. Zhang and J. Gao were with equal contributions.
Y. Wang and W. Tang were the corresponding authors.\protect\\
E-mail: \{malt, wangyasha\}@pku.edu.cn; tanggwen@126.com\protect
\IEEEcompsocthanksitem J. Gao was with Health Data Research and University of Edinburgh, UK.
\IEEEcompsocthanksitem W. Tang and T. Wang were with Department of Nephrology, Peking University Third Hospital, CN.
\IEEEcompsocthanksitem X. Zhao was with Department of Nephrology, Peking University People's Hospital, CN.
\IEEEcompsocthanksitem W.Ruan was with the Department of Computer Science, University of Exeter, UK.
}
}

\markboth{2023}%
{Ma \MakeLowercase{\textit{et al.}}: Mortality Prediction with Adaptive Feature Importance Recalibration for Peritoneal Dialysis Patients}

\IEEEtitleabstractindextext{%
\begin{abstract}


\textbf{Objective:}
Peritoneal Dialysis (PD) is one of the most widely used life-supporting therapies for patients with End-Stage Renal Disease (ESRD). 
Predicting mortality risk and identifying modifiable risk factors based on the Electronic Medical Records (EMR) collected along with the follow-up visits are of great importance for personalized medicine and early intervention.
Recent studies have attempted to utilize machine learning techniques to evaluate the health status of patients.
However, there are still some critical issues that have not yet been thoroughly addressed by existing work in terms of 
\textit{1)} effective utilization of both sequential medical records and baseline demographic information,
\textit{2)} providing fine-grained interpretability by selecting key features for each patient individually, and
\textit{3)} adaptively analyzing the changes of feature importance with its value and extracting related medical knowledge.
Here, our objective is to develop a deep learning model for a real-time, individualized, and interpretable mortality prediction model -- \mname.

\textbf{Method and Materials:} 
Our proposed end-to-end model consists of a multi-channel feature extraction module and an adaptive feature importance recalibration module. 
Both dynamic records and static features are embedded together to form a squeezed health context representation and further guide the assignment of the attention weights.
\mname explicitly identifies the key features that strongly indicate the outcome prediction for each patient to build the health status embedding individually.
This study has collected 13,091 clinical follow-up visits and demographic data of 656 PD patients from Peking University Third Hospital, covering more than 12 years.
There are about 20 visits recorded for each patient, with an average visit interval of 2.7 months and an average follow-up time of 4 years.
To verify the application universality, this study has also collected 4,789 visits of 1,363 hemodialysis dialysis (HD) as an additional experiment dataset to test the prediction performance, which will be discussed in the Appendix.

\textbf{Results and Conclusion:}
To summarize:
\textit{1)} 
Experiment results show that \mname achieves 81.6\%/74.3\% AUROC (area under the receiver operating characteristic curve) and 47.2\%/32.5\% AUPRC (area under the precision-recall curve) for the 1-year mortality prediction task on PD/HD dataset respectively, which outperforms the state-of-the-art comparative deep learning models while simultaneously providing qualitative interpretability.
\textit{2)} This study first provides a comprehensive elucidation of the relationship between the causes of mortality in patients with PD and clinical features based on an end-to-end deep learning model. 
\textit{3)} This study first reveals the pattern of variation in the importance of each feature in the mortality prediction and provides recommended reference value for most PD patients based on built-in interpretability, without any injection of human physicians' knowledge. 
\textit{4)} To further foster personalized clinical service, we develop a practical AI-Doctor interaction system to visualize the trajectory of patients' health status and risk indicators.
The prototype of the system has already been on trial via experienced nephrologists.

\textbf{AI-Doctor Interaction Online System:} 
We have made an abstract presentation video introducing our work to aid your review \url{https://youtu.be/CY2glHchsC8}.
Our developed health trajectory visualization system with anonymous case studies (patient ID from A1 to A20) is publicly available at \url{http://47.93.42.104/A8}. 
Visualization of the importance of the features is available at \url{http://47.93.42.104/statistics/feature}. 
We release our code at \url{https://github.com/Accountable-Machine-Intelligence/AICare}.
Users can upload the data online to get the prediction results immediately \url{http://47.93.42.104/predict} or download the code to train the model based on their dataset offline.

\end{abstract}

\begin{IEEEkeywords}
Electronic Medical Record (EMR); End-Stage Renal Disease (ESRD); Peritoneal Dialysis (PD); Mortality Prediction; Deep Learning; Model Interpretability.
\end{IEEEkeywords}}

\maketitle
\IEEEdisplaynontitleabstractindextext
\IEEEpeerreviewmaketitle

\section{Introduction}

The prevalence of End-Stage Renal Disease (ESRD) continues to increase and has become a significant healthcare burden worldwide. 
Approximately 3.8 million people currently rely on some form of dialysis for the treatment of ESRD worldwide \cite{teitelbaum2021peritoneal}.
ESRD is a long-term disease, and patients need continuous medical care and treatment for years or even decades.
Peritoneal Dialysis (PD) is a well-established Renal Replacement Therapy (RRT) modality and the leading form of home-based life-supporting dialysis therapy for patients with ESRD \cite{morelle2021aqp1}. 
Over the past decade, the use of peritoneal dialysis increased dramatically worldwide.


During the long-term peritoneal dialysis, patients may still suffer various vital risks such as cardio-cerebrovascular disease and infection \cite{bender2006prevention}. 
These risks may cause adverse outcomes, and patients need lifelong treatment with periodic follow-up visits to monitor their health status. 
Predicting mortality risk and identifying modifiable risk factors from routine clinic visit records are of great importance for personalized medicine and early intervention to prevent adverse outcomes and improve the survival of long-term PD patients. 
Recent studies have attempted to utilize Artificial Intelligence (AI) techniques to evaluate the health status of patients.
However, as shown in the Summary Table~\ref{tab:relatedwork} of Related Works in the Appendix section \ref{sec:related_work}, there are still some critical issues that have not yet been thoroughly addressed by existing works in terms of:


\textbf{$I_1$: Perform dynamic mortality prediction at each follow-up visit based on the effective utilization of both sequential medical records and the baseline demographic information.} 
Most AI-based EMR analysis research on kidney disease patients only use static baseline information to perform one-time health prediction based on traditional machine learning methods \cite{xu2019stratified,ravizza2019predicting,chaudhuri2021machine,akbilgic2019machine,liu2021predicting,zhou2021prediction,radovic2022machine,kang2020machine,noh2020prediction,schena2021development}.
These methods cannot perform real-time health prediction, and thus the practical utility in the clinical application is limited.
Some other research models the disease process by incorporating sequential EMR \cite{rank2020deep,tomavsev2019clinically,srinivas2017big}.
However, these works cannot yet effectively embed the baseline information and the sequential records together, and capture the interaction between them during the health status embedding procedure, which leads to limited prediction performance. 

\textbf{$I_2$: Provide fine-grained interpretability for each patient individually by selecting key features which contribute the most to mortality prediction (patient-level interpretability) and achieve high prediction performance simultaneously.}
Key factors that strongly indicate health risk are different among patients. 
Medical experts need to understand how a model makes a specific decision for a particular patient.
This requires sufficient model interpretability to ensure that prediction results are trustworthy for developing bespoke interventions and extracting medical knowledge. 
However, most existing works fail to ensure the model's trustworthiness in providing verifiable interpretations.
On the one hand, traditional machine learning models, such as decision trees \cite{noh2020prediction,akbilgic2019machine,liu2021predicting,kang2020machine,yan2020interpretable}, are clinically interpretable, but they cannot capture complex longitudinal progressions and thus have inferior prediction performances.
On the other hand, the decision-making process in deep-learning-based models is a black box and fails to provide human-understandable interpretation \cite{rank2020deep,schena2021development}. 
Although some recent works apply the SHapley Additive exPlanations (SHAP) \cite{hyland2020early,thorsen2020dynamic}, feature permutation \cite{zhou2021prediction,radovic2022machine}, and inverse analysis \cite{makino2019artificial} strategies to improve the interpretability.
However, these post-hoc interpreting \cite{alvarez2018towards} methods can only provide coarse-grained interpretability, which is difficult to understand at the patient level.
It is still challenging to provide satisfactory interpretability and achieve high prediction performance simultaneously. 

\textbf{$I_3$: Adaptively analyze the importance of each feature along with the variation of its value (feature-level interpretability) to provide medical advice and extract knowledge.}
The way of attending to the medical feature in the prediction process should be flexible and individualized according to its value.
However, most existing works analyze the health status of patients in a fixed decision process \cite{noh2020prediction,akbilgic2019machine,yan2020interpretable} or embed clinical features via fixed parameters of neural networks without ante-hoc interpretability  \cite{zhou2021prediction,schena2021development,makino2019artificial}.
To the best of our knowledge, none of the existing AI-based clinical prediction work for kidney disease patients explicitly analyzes the changes in the feature importance with features' values.


\begin{figure*}[]
  \centering
  \includegraphics[width=1.75\columnwidth,trim={0cm 0 0cm 0cm},clip]{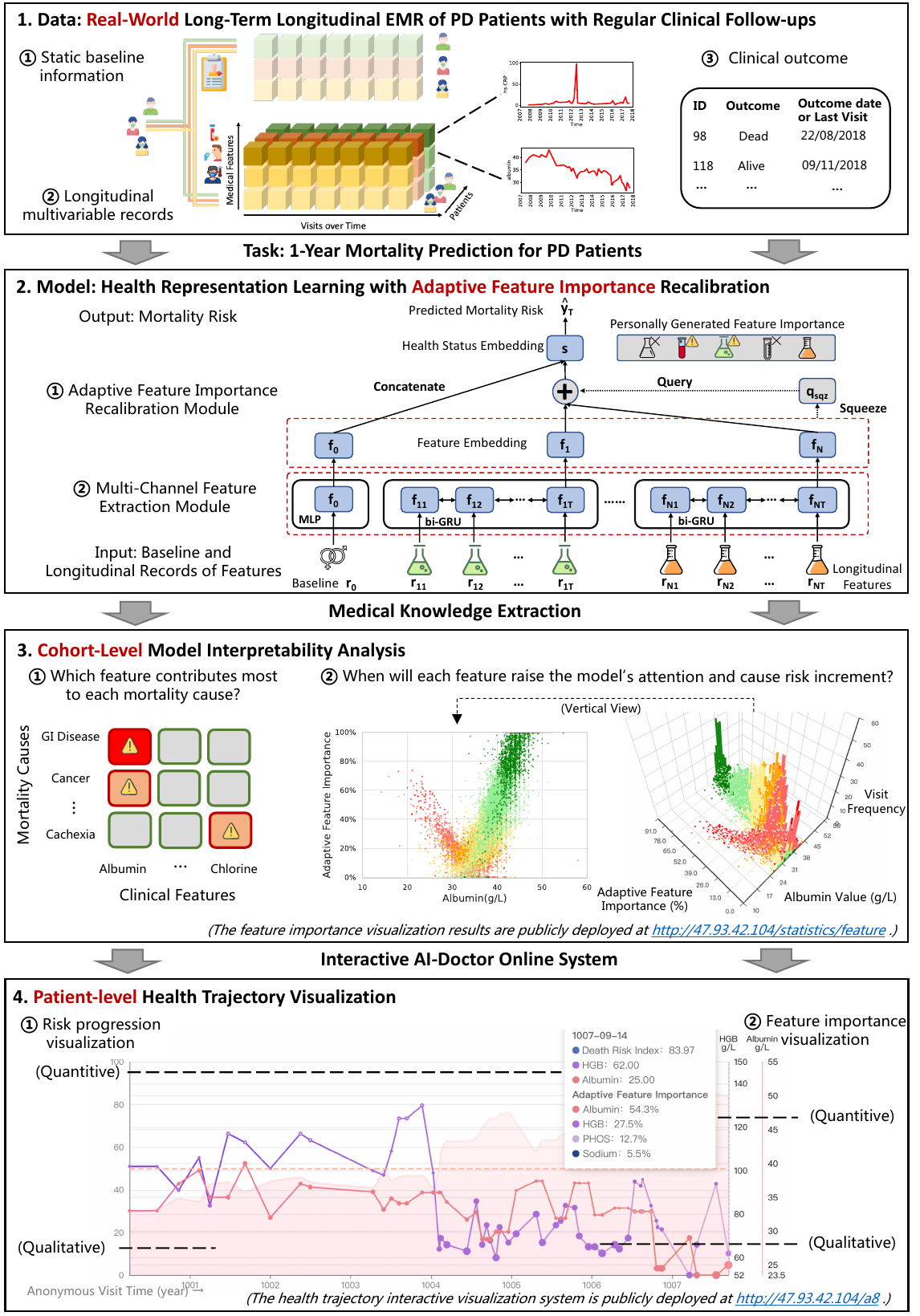}
  \caption{\textbf{Mortality Prediction Research Overview for Peritoneal Dialysis (PD) Patients.}
  \textit{1)} We collect an over {\it 12-year}, {\it long-term}, and {\it real-world} clinical EMR dataset of PD patients, consisting of static baseline information, longitudinal multi-variable records and clinical outcomes. The prediction task is defined as a 1-year mortality prediction at each clinical visit.  
  \textit{2)} We propose a deep-learning-based interpretable health status representation learning framework, consisting of a multi-channel feature extraction module and an adaptive feature importance recalibration module. 
  \textit{3)} We perform the model interpretability analysis for diverse mortality causes and observe the change of feature importance to extract novel medical knowledge (taking albumin as an example). 
  \textit{4)} We build an interactive AI-Doctor system to visualize the health trajectory.}
  \label{fig:overview}
\end{figure*}

To address the above challenges, we propose a deep-learning-based interpretable mortality risk prediction framework for PD patients, \mname. 
As shown in Fig.~\ref{fig:overview}-(1), it is built upon a real-world longitudinal EMR dataset of PD patients spanning over 12 years, including baseline demographic information and outcomes, as well as patient-level follow-up lab tests and diagnosis records spanned by an average of 20 visits per patient. 
The main contributions of this work are summarized below:

\begin{itemize}[leftmargin=*]

\item Our proposed framework, \mname, models the health trajectory based on multivariate EMR data of PD patients and achieves better prediction performance than state-of-the-art (SOTA) methods while simultaneously providing fine-grained patient-level interpretability.


\begin{itemize}[leftmargin=*]

\item As shown in Fig.~\ref{fig:overview}-(2), \mname employs a multi-channel medical feature embedding architecture to extract sequential patterns from high-dimensional medical features. 
The squeezed embedding of static information and dynamic features is treated as a health context vector to guide the feature importance recalibration (addressing $I_1$).
\mname assigns attention weights for each feature by looking at the health context for clues that can help lead to a more individual representation of the health status.

\item On the dynamic mortality prediction tasks, \mname achieves 47.2\% AUPRC on the PD dataset, which is 11.8\% relatively higher than the SOTA comparative baseline model.
We also introduce an additional experiment dataset, which will be described in the Appendix.

\end{itemize}

\item \mname provides the first comprehensive elucidation of the relationship between the causes of mortality in patients with PD and clinical features (patient-level interpretability in $I_2$) based on an end-to-end deep learning model.
As shown in Fig.~\ref{fig:overview}-(3.1) and Fig.~\ref{fig:rawattention}, \mname achieves fine-grained interpretability by explicitly and adaptively emphasizing high-risk features during the prediction process based on a feature recalibration module.
The interpretation from \mname is not only consistent with existing medical research but also helps clinicians make reliable clinical decisions and facilitates the discovery of new medical knowledge. 
We report detailed patient-level interpretability analyses, such as:

\begin{itemize}[leftmargin=*]

\item Serum albumin, diastolic blood pressure, and chlorine are the most important indicators for most PD patients. 
\item Albumin has a strong indication for patients who died of gastrointestinal disease and peripheral vascular disease. 
\item Diastolic blood pressure (DBP) has an indication for patients who died of cachexia, cancer, and cerebrovascular disease. 
Systolic Blood Pressure (SBP) is indicative of cancer and peritoneal dialysis-associated peritonitis deaths.
\end{itemize}

\item To the best of our knowledge, \mname first reveals the variation pattern in each feature's importance for PD patient's mortality prediction (feature-level interpretability in $I_3$), without any injection of human physicians' knowledge.
As shown in Fig.~\ref{fig:overview}-(3.2), Fig.~\ref{fig:feature-importance-variation} and Table~\ref{tab:turningpoint}, 
\mname provides the ante-hoc attention weight of each clinical feature according to its value and the patient's condition.
We report detailed feature-level interpretability analyses, such as:

\begin{itemize}[leftmargin=*]

\item There are two variation patterns of importance in medical features, \textit{V-shaped} parabolic curves (e.g., albumin, diastolic blood pressure) and \textit{L-shaped} fold lines (e.g., systolic blood pressure, hemoglobin). 

\item For example, the importance weight of albumin is presented as a V-shaped curve with 32 g/L as the lowest turning point.
For most PD patients, when albumin is lower (higher) than the turning point of 32 g/L, the more extreme the value is, the more attention weight is assigned by \mname, which means that this feature plays an essential role in the health status representation learning, and the predicted mortality risk rises (declines). 
\mname recommends improving albumin to higher than 32 g/L, the higher the better.

\item The importance weight of SBP presents as an L-shaped curve with 130 mmHg as a turning point.
For SBP level over 130 mmHg, \mname pays nearly no attention to SBP.
\mname recommends raising the SBP level to at least 130 mmHg for most PD patients, but the further increment of SBP will not bring many benefits. 

\end{itemize}

\item We develop a practical AI-Doctor interaction system to visualize the trajectory of patients' health status and risk indicators.
Model deployment has been the last but most challenging step toward clinical applications.
Deploying the deep models in an accessible way for clinicians to allow them to easily understand model predictions and model decision process still needs extra considerations. 
As shown in Fig.~\ref{fig:overview}-(4), to further facilitate personalized clinical service, we deploy an AI-Doctor interaction system online with open-source code at \url{https://github.com/Accountable-Machine-Intelligence/AICare}.

\end{itemize}

\section{Results}

\subsection{Data Description and Problem Formulation}

We have collected the EMR data of 656 PD patients with 13,091 visit records, spanning over 12 years, from January 1, 2006, to January 1, 2018, including patients' baseline data, longitudinal visit records and outcomes.
\begin{itemize}
    \item \textbf{Baseline data} include patients' demographic data (e.g., age, gender) and the diagnosis of diabetes at the beginning of dialysis. 
    The statistics of the baseline data and the assignment of the labels are shown in Table~\ref{tab:statistics}.
    \item \textbf{Visiting data} includes laboratory tests and patients' vital signs at each visit. 
    The visiting frequency statistics and the feature values distribution are shown in Table~\ref{tab:range} and Table~\ref{tab:feature_summary}, respectively. 
    \item \textbf{Outcome data} includes patients' outcomes at the end of the data collection window, including death date and causes of death (e.g., cancer). 
    The outcomes of all patients were followed and further recorded until October 31, 2018.
\end{itemize}


The feature sets consist of 16 longitudinal medical features and 4 baseline features. 
The age distribution of the patients is 58.55 $\pm$ 15.81 years, and the number of average records per patient is 19.95 $\pm$ 13.53. 
We fill in the missing values with the most recent historical recorded values.

We conduct the \textbf{one-year dynamic mortality prediction} task.
Given a patient's visit records with $T$ visits, the binary classification task is to predict the mortality risk in the future one year $\hat{y}_t$ at each visit $t$. 
In order to meet the actual clinical practice, we also define an uncertain phase of patients' health status. 
For patients with negative labels (alive, $y=0$), the uncertain phase is one year
before the end date of data collection, since we do not know the outcomes of these patients in the future one year. 
For patients with positive labels (dead, $y=1$) at $t$, the uncertain phase is between the $t-2$
year and the $t-1$ year, since we are uncertain about the ground-truth health status during these visits.  
The final dataset contains 1,196 visits with positive labels (i.e., died within one year) and 10,804 records with negative labels.
For more details about the dataset and the problem formulation, please see the Appendix section \ref{sec:dataset} and \ref{sec:problem_formulation}, respectively.

\begin{table}
\centering
  \caption{\textbf{Statistics of Baseline Information and Label Assignment.} 
  The real-world dataset contains 656 Peritoneal Dialysis (PD) patients with 13,091 clinical visits. 
  There are 39.8\% patients, unfortunately, who died before the final follow-up. 
  The age range of patients enrolled is from 16 to 98 years old.  
}
  \label{tab:statistics}
  \begin{tabular}{lccc}
     \hline
     & Total & Mortality (\%) & Survival (\%) \\
     \hline
    \# Patients & 656 & 261 (39.8\%) & 395(60.2\%)\\
    \# Visits & 13091 & 1196 (9.1\%)& 11895 (90.9\%)\\
    \hline
    \textbf{Age} & & & \\
    \hline
    16-40 & 96 (14.6\%) & 10 (10.4\%) & 86 (89.6\%)\\
    40-60 & 217 (33.1\%) & 64 (29.5\%) & 153 (70.5\%) \\
    60-80 & 297 (45.3\%) & 153 (51.5\%) & 144 (48.5\%)\\
    80-98 & 44 (6.7\%)& 33 (75.0\%) & 11 (25.0\%)\\
    \hline
    \textbf{Diabetes} & & & \\
    \hline
    \# Diabetes & 244 (37.2\%) & 120 (49.2\%) & 124 (50.8\%) \\
    \hline
    \textbf{Gender} & & & \\
    \hline
    \# Female & 327 (49.8\%) & 125 (38.2\%)&202 (61.8\%)\\
    \# Male & 328 (50.2\%) & 136 (41.5\%)&192 (58.5\%)\\
    \hline
\end{tabular}
\end{table}

\begin{table}[]
\centering
  \caption{\textbf{Statistics of Age and Visiting Frequency.} 
  Peritoneal Dialysis (PD) patients were followed up every 3 months. 
  There are about 20 visits recorded for each patient.  }
  \label{tab:range}
  \begin{tabular}{cccccc} 
     \hline
    Statistic & Avg. &Med. & Max. & Min. & Std.\\
     \hline
     Age (year) & 58.55 &60.70 & 97.45 & 16.79 &15.81 \\
    Visits per Patient& 19.95 &16 & 69 & 1 &13.53 \\
    High Risk Visits per Patient & 2 &0 & 29 & 0 &2.95 \\
    Duration of Follow-up (year) & 3.98 & 3.43 & 10.44 & 0.1 & 2.67 \\
    Visit Interval (month) & 2.73 & 2.48 & 29.87 & - & 2.67 \\
   \hline
\end{tabular}
\end{table}

\begin{table*}[]
\small
 \centering
  \caption{\textbf{Feature Summary of Peritoneal Dialysis (PD) Dataset.} 
  This dataset comprises 16 dynamic features recorded at each clinical visit and 4 static baseline features recorded at the first visit.}
  \label{tab:feature_summary}
  \begin{tabular}{ccccccccccc}
\hline
     & Abbreviation & Full Name & Unit & \multicolumn{3}{c}{High Risk Visits ($y=1$)}   & \multicolumn{3}{c}{Low Risk Visits ($y=0$)} & \% Missing  \\
\hline
     & \multicolumn{2}{l}{Dynamic Features}   & & Mean & Std & Median & Mean & Std &Median &\\
\hline
    & Albumin &  Albumin  &g/L & 33.81 &4.437 &34.3 &37.87 &4.337 &38 & 25\% \\
    & DBP   &  Diastolic Blood Pressure &mmHg& 70.28 &14.71 & 70& 78.59& 13.79& 80& 18\%\\
    & SBP &  Systolic Blood Pressure & mmHg&125.3 &25.19& 127&134.4 & 21.61& 135& 14\%\\
    & Cl &  Chlorine & mmol/L & 96.02 &4.155&96 & 98.21&4.923 & 98&17\% \\
    & Cr &  Creatinine & umol/L & 779.6 & 250.3 & 741&868.9 & 270.3 &853 &10\%\\
    & Urea &  Urea & mmol/L & 18.12 &5.545 & 17.8& 20.09& 5.363 &19.8&11\%  \\
    & Ca &  Calcium & mmol/L & 2.358 & 0.277 & 2.345& 2.406 &0.341 &2.39 &12\%\\
    & Na &  Sodium& mmol/L & 137.1 & 4.262 & 137.9 & 138.5& 4.617  &139 &21\%  \\
    & K &  Potassium & mmol/L & 4.240& 0.783 & 4.17& 4.320& 0.718 & 4.25 &11\%\\
    & P &  Phosphorus& mmol/L& 1.549 & 0.450 &1.5 &1.606 &0.430 &1.57 &13\%  \\
    & CO$_2$CP &  CO$_2$ Combining Power& mmol/L& 27.45& 3.562 & 27.5 &27.38 & 3.630 &  27.4 &8\%\\
    & Hb &  Hemoglobin &g/L & 111.4 & 19.54 & 113 & 114.6&17.05 & 115&12\%\\
    & Weight &  Body Weight &kg & 59.98 & 11.05 &59.59 & 62.26& 11.07 & 62&41\%\\
     & Glucose &  Glucose& mmol/L & 7.758 & 3.665 & 6.7&6.689 & 3.089 & 5.7& 30\% \\
    & hs-CRP &  Hypersensitive C-Reactive Protein &mg/L & 17.57 & 28.07 & 8.49& 7.954& 13.96 & 3.19 &29\%\\
    & WBC &  White Blood Cell Count& x10$^9$/L & 8.238 & 2.767 &7.895 &7.773 &2.754 &7.43& 10\%\\
\hline
    & \multicolumn{2}{l}{Baseline Features}    & &  &  &  &  &  & & \\
\hline
     & Age &  Age & year & 66.12 & 13.01 & 67.82 & 53.30 & 15.54 & 54.53 & 0\% \\
     & Gender &  Female (0) or Male (1) & - & 0.53 & 0.50 & 1 & 0.49 & 0.50 & 0 & 0\%\\
     & Height &  Height& m & 162.2 & 9.95 & 160.5 & 164.1 & 10.98 & 163.8 & 0\%\\
      & Diabetes &  Is (1) or Not (0) Has Diabetes &- & 0.45 & 0.50 & 0 & 0.31 & 0.46 & 0 & 0\%\\
\hline
\end{tabular}
\end{table*}

\subsection{Prediction Performance}

The prediction performance of \mname and the baseline models on the 10-fold cross-validation mortality prediction of PD patients are shown in Table~\ref{tab:performanceAll}.
\mname achieves 47.2\% AUPRC, which is relatively 11.8\% higher than the best baseline model \footnote{${}^{**}:p<0.01$; ${}^{*}:p<0.05$}. 
This indicates that \mname can efficiently embed the long-term longitudinal multi-variable sequential data and static baseline data to learn the representation of the health status of PD patients individually,
using the multi-channel feature extraction module and the adaptive feature importance recalibration module.
More details about the experiment are listed in the Appendix sections, including:
\textit{1)} The prediction performance of diverse mortality causes is listed in Fig.~\ref{fig:results-cod}.
\textit{2)} The detailed descriptions of the comparative baseline methods are listed in Appendix section \ref{sec:cod}.
\textit{3)} To verify the application universality of \mname on other patient cohorts, we also introduce an additional experiment dataset to train the model and test the prediction performance, which consists of 1,363 ESRD patients receiving Hemodialysis Dialysis (HD) from Peking University People's Hospital.
The prediction performance is listed in Table~\ref{tab:performance-external}.



\begin{table}[]
\centering
\caption{\textbf{Mortality Prediction Performance of PD Patients.} Our proposed model, \mname, outperforms other baseline comparative approaches, including deep models. The values in parentheses () are the standard deviation of 10-fold cross-validation. }
    \label{tab:performanceAll}
\begin{tabular}{ccccccc}
\hline
Method     & AUPRC & AUROC \\
\hline
GRU~\cite{meyer2018machine}      &  0.422\,(0.109)&  0.781\,(0.047) \\ 
Transformer~\cite{nitski2021long}   & 0.406\,(0.097) & 0.789\,(0.047) \\ 
MT-RHN~\cite{tomavsev2019clinically} & 0.413(0.107) & 0.777(0.063) \\ 
LSTM~\cite{thorsen2020dynamic}                & 0.395(0.100) & 0.782(0.065) \\ 
biLSTM-FC~\cite{sung2021event}              & 0.398(0.089) & 0.758(0.067) \\ 
LR~\cite{ravizza2019predicting}   & 0.370\,(0.084) & 0.610\,(0.044) \\ 
XGBoost~\cite{yan2020interpretable}             & 0.379\,(0.087) & 0.597\,(0.033) \\ 
DT~\cite{noh2020prediction}       & 0.319\,(0.040) & 0.607\,(0.027) \\ 
LightGBM~\cite{hyland2020early}            & 0.405\,(0.082) & 0.604\,(0.028) \\ 
\hline
\mname  & \textbf{0.472}$^{**}$\,(0.075)& \textbf{0.816}$^{**}$\,(0.033) \\
\hline
\end{tabular}
\end{table}

\begin{figure}[]
  \centering
  \includegraphics[width=1.1\columnwidth]{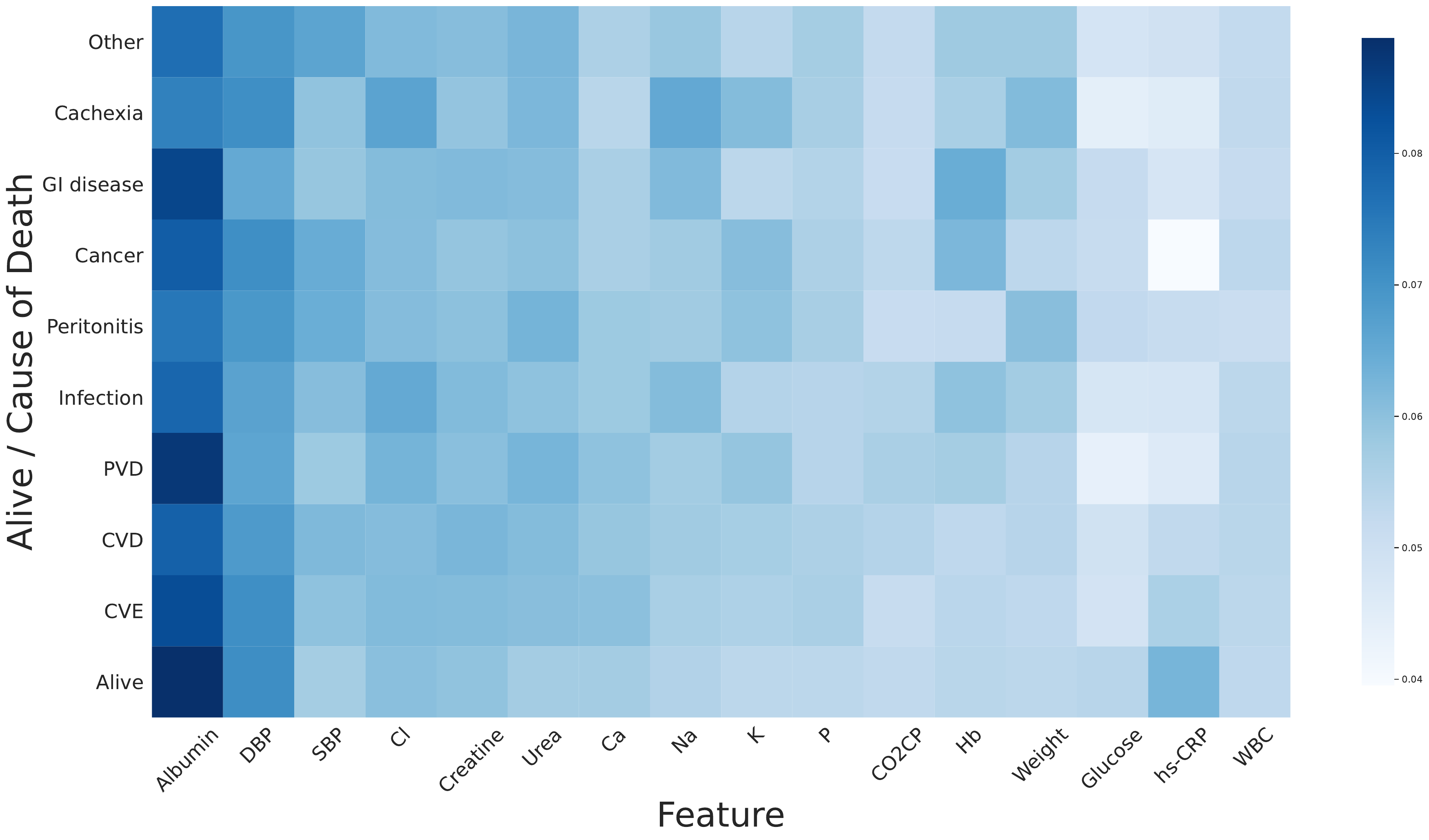}
  \caption{\textbf{Average Feature Importance Heatmap for Diverse Mortality Causes Generated by \mname.} 
  The \textbf{\textcolor{blue}{darker}} the color, the greater the importance. 
  \textit{1)} Serum albumin is the most important feature in mortality prediction, especially for Peritoneal Dialysis (PD) patients who died of gastrointestinal (GI) disease and peripheral vascular disease (PVD). 
  \textit{2)} Diastolic blood pressure (DBP) is the second indicative feature, especially for PD patients who died of cachexia, cancer and cerebrovascular disease (CVE).}
  \label{fig:rawattention}
\end{figure}

\subsection{Interpretability Analysis}

\mname provides fine-granular interpretability to help clinicians understand the prediction decision process. 
At each visit, the model provides dynamic importance weights for the features, which indicate the contributions of each feature to the final prediction result. 
In this section, we discuss the detailed interpretability analyses.

\subsubsection{Average feature importance for diverse causes of death}

We calculate the average importance of each feature for patients, which is shown as a heatmap in Fig.~\ref{fig:rawattention}. 
The results indicate that serum \textit{albumin, diastolic blood pressure (DBP)}, and \textit{chlorine (Cl)} are considered as important health indicators for most PD patients as their columns are darker than other feature columns. 
Some findings of the relationship between the causes of mortality in PD patients and clinical features are listed below:

\begin{itemize}[leftmargin=*]

\item \textbf{Albumin} is the strongest indicator of most causes of death, especially for \textit{cerebrovascular diseases (CVE), peripheral vascular disease (PVD)} and \textit{gastrointestinal (GI) disease}, according to the heatmap generated by \mname. 
This may be because albumin is an indicator of protein-energy wasting, correlated with suboptimal gastrointestinal intake and inflammation \cite{tsirpanlis2005serum,de2009association}. 
Hypoalbuminemia is a strong predictor for PD-related peritonitis \cite{li2016ispd}, which is the primary reason for deaths of infection and peritonitis. 
Besides, our model generates a high attention weight of albumin for patients who are still \textit{alive}, which means that low-risk scores are associated with high albumin value.
More details about albumin will be discussed in subsection \ref{sec:turningpoint}.

\item \textbf{Diastolic blood pressure (DBP)} is a risk indicator for \textit{CVE, PD-related peritonitis, cancer} and \textit{cachexia} deaths.
This may be because DBP level is a marker of atherosclerosis and is strongly independently related to atherothrombotic brain infarction incidence \cite{jeerakathil2003epidemiology}. 
Low DBP could also be an indicator for low peripheral vascular resistance or increased arterial stiffness \cite{o1998mcdonald, fang1995measures} which is strongly associated with a high incidence of cardio-cerebral vascular disease \cite{webb2020progression}.
Additionally, low BP is a surrogate predictor for specific comorbidities, heart failure, chronic inflammation and malnutrition \cite{zager2009blood}, which may relate to death from peritonitis, cancer and cachexia. 

\item \textbf{Sodium (Na), potassium (K)} and \textbf{body weight} are important indicators for \textit{cachexia} deaths.
This may be because patients with cachexia often experience low sodium and potassium level due to insufficient low food intake. 
Decreasing in weight for these patients is a common phenomenon. 

\item \textbf{Hemoglobin (Hb)} is an important indicator for \textit{Gastrointestinal (GI) disease} deaths. 
GI bleeding is a critical manifestation of uremic GI disease. 
\textbf{Hemoglobin (Hb)} and \textbf{potassium (K)} are indicators for \textit{cancer} deaths, which are consistent with the fact that cancer is highly associated with refractory anemia, anorexia, and, consequently, hypokalemia due to insufficient intake.

\item \textbf{Urea, body weight, potassium (K), albumin, diastolic blood pressure (DBP)} and \textbf{systolic blood pressure (SBP)} are important indicators for \textit{PD-related peritonitis} deaths.
The risk factors for \textit{peritonitis}, a common complication of PD patients, have been well-defined \cite{li2016ispd}, including hypoalbuminemia, hypokalemia, protein energy-wasting, etc.
This is consistent with the results of our model.
\end{itemize}

\begin{figure*}[]
  \centering
  \includegraphics[width=2.15\columnwidth]{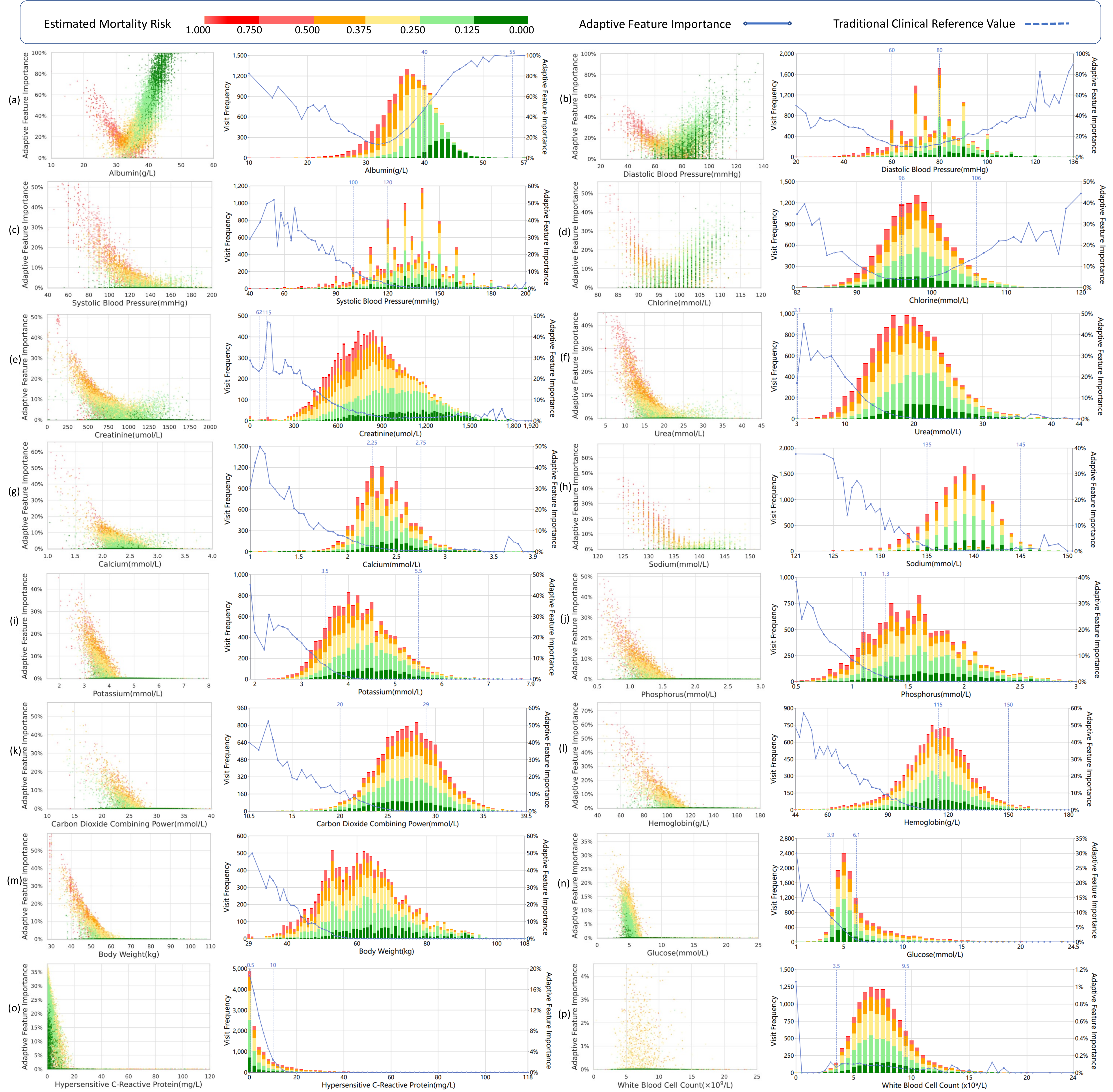}
  \caption{\textbf{Feature Importance Variation Learned by \mname.}~\textit{(Best viewed in color.)} 
  The clinical visits are marked as colored dots and histograms. 
  \textcolor{red}{Red} represents \textcolor{red}{high risk} predicted by \mname, while \textcolor{green}{green} represents \textcolor{green}{low risk}. 
  The average feature importance is visualized as \textcolor{blue}{blue} fold lines (\textcolor{blue}{——}). 
  The traditional clinical reference values are vertically marked as \textcolor{blue}{blue} dotted lines (\textcolor{blue}{:}). 
  There are two variation patterns of feature importance, V-shaped parabolic curves (e.g., albumin, diastolic blood pressure) and L-shaped fold lines (e.g., systolic blood pressure, hemoglobin). 
  \textbf{\textit{1)}} We take the serum albumin's importance variation as an example of a \textbf{V-shaped} parabolic curve. 
  For most patients, when albumin is lower (higher) than the turning point of 32 g/L, the more extreme the value is, the more attention weight is assigned by \mname, which means this feature takes essential parts in health status representation learning, and meanwhile the predicted mortality risk rises (declines). 
  As a result, \mname recommends improving the serum albumin to above 32 g/L, the higher the better.
  \textbf{\textit{2)}} On the contrary, we take the systolic blood pressure's (SBP) importance variation as an example of an \textbf{L-shaped} fold line. 
  For most patients, when SBP is lower than the turning point of 130 mmHg, the lower the value is, the more attention weight is assigned. 
  However, when SBP is higher than 130 mmHg, the attention weight drops to nearly 0\%, meaning this feature will no longer affect the representation learning of health status. 
  As a result, \mname recommends improving the SBP to at least 130 mmHg, but higher will not bring many benefits.
  The quantified summary is listed in Table~\ref{tab:turningpoint}.
  The visualization results are publicly deployed at \url{http://47.93.42.104/statistics/feature}.}
  \label{fig:feature-importance-variation}
\end{figure*}

\subsubsection{Change of feature importance with feature values}
\label{sec:turningpoint}
\begin{table*}[]
\centering
\caption{\textbf{Importance Variation Pattern and Recommended Reference Value (Turning Point) Learned by \mname for PD Patients.} 
This table is a quantified summary of Fig.~\ref{fig:feature-importance-variation}.
Recommendation "\textit{Higher}" means that \mname suggests increasing this feature's value above the turning point; 
"\textit{At Least}" means that \mname suggests maintaining the value above the turning point, but further increment may not bring many benefits.
Consistency "$\sim$" means that there is some overlap between the reference range recommended by \mname for PD patients and the traditional reference range for outpatients.
We have publicly deployed the visualization charts of the variation in importance of features via \url{http://47.93.42.104/statistics/feature}. }
\label{tab:turningpoint}
\begin{tabular}{lc|ccc|cc|c}
\hline

Feature & Unit & \multicolumn{3}{c|}{Importance Variation Learned by \mname} & \multicolumn{2}{c|}{Traditional Reference Range} & Consistency  \\
 & & Variation Type & Recommendation & Turning Point & Lower Limit & Upper Limit &   \\
\hline

Albumin &g/L& \textbf{V-Shape} & \textbf{Higher} & \textbf{\textgreater 32}  & 40 & 55  & $\surd$ \\
DBP&mmHg&\textbf{V-Shape}& \textbf{Higher} & \textbf{\textgreater 70}  & 60  & 80 & $\sim$ \\
SBP&mmHg& L-Shape & At Least& \textbf{\textgreater 130} & 100  & 120 & $\times$  \\
Chlorine& mmol/L& \textbf{V-Shape}& \textbf{Higher} & \textbf{\textgreater 96} & 96 & 106  &$\surd$ \\
Creatinine& umol/L&L-Shape& At Least & \textbf{\textgreater 900} & 62 & 115 & $\times$  \\
Urea& mmol/L&L-Shape & At Least& \textbf{\textgreater 20}& 3.1 & 9 & $\times$  \\
Calcium& mmol/L&L-Shape & At Least& \textbf{\textgreater 2.5} & 2.25 & 2.75 & $\sim$  \\
Sodium& mmol/L&L-Shape &At Least&\textbf{\textgreater 135.5} & 135 & 145 &$\surd$ \\
Potassium& mmol/L& L-Shape&At Least&\textbf{\textgreater 4} & 3.5 & 5.5 & $\surd$\\
Phosphorus& mmol/L& L-Shape&At Least& \textbf{\textgreater 1.5} & 1.1 & 1.3 & $\times$  \\
CO2CP& mmol/L&L-Shape &At Least& \textbf{\textgreater 25} & 20 & 29 & $\sim$  \\
Hemoglobin& g/L& L-Shape& At Least& \textbf{\textgreater 114} & 115 & 150 & $\surd$\\
Weight&kg&L-Shape &At Least& \textbf{\textgreater 59}& - & -  &-\\
Glucose& mmol/L& L-Shape&Not Exceed &\textbf{\textless 6} & 3.9 & 6.1  &$\surd$\\
Hs-CRP&mg/L&L-Shape &Not Exceed & \textbf{\textless 16} & 0.5 & 10 & $\surd$\\
WBC& x10$^9$/L&Irregular & Unknown & - & 3.5 & 9.5 & -\\

\hline
\end{tabular}
\end{table*}

\mname quantifies the feature importance changes with feature values in a macroscopic perspective for the whole patient cohort to help clinicians better understand the decision process, take an individualized intervention, and extract new medical knowledge, as shown in Fig.~\ref{fig:feature-importance-variation}.
\begin{itemize}[leftmargin=*]
\item In the left scatter plot, the $x$-axis denotes the value of the biomarker.
The $y$-axis denotes the feature's importance.
Each dot represents a follow-up visit of a patient, and the color represents the predicted risk. 

\item The right histogram shows the risk distributions at different values of biomarkers.
The blue curve is the fitted curve of the average importance of the feature. 
We also plot each feature's traditional clinical reference ranges for normal clinic outpatients as blue dotted lines, helping physicians evaluate the consistency between the results of \mname and the traditional ranges. 
\end{itemize}

There are two obvious patterns of relationships between biomarkers' importance weights and recorded values: V-shaped parabolic curve and L-shaped fold line. 
\begin{itemize}[leftmargin=*]
\item For the \textit{V-shaped} parabolic pattern (e.g., albumin, diastolic blood pressure), an extremely high or low feature value will cause high importance attention weight through \mname, which means the feature plays an essential part in the learning of the representation of health status. 

\item For the \textit{L-shaped} fold line pattern (e.g., systolic blood pressure, hemoglobin), the lower the biomarker value, the higher the importance of attention weight. 
\end{itemize}

The pattern of variation in importance and the recommended reference values learned by \mname are summarized in Table~\ref{tab:turningpoint}.
We will discuss these patterns in detail in the following text. 

\textbf{1) Albumin} Fig.~\ref{fig:feature-importance-variation}-(a). 
\mname believes that the albumin's importance attention weight appears to be a V-shaped curve with 32 g/L as a turning point.
The variation of albumin in a descending or ascending manner always gets the model's attention.
Considering that the red dots are mostly on the left side of the figure, \mname learns that patients with albumin-level lower than 32 g/L tend to have a high importance weight and poor prognosis ($y>0.5$). 
When the albumin level is lower than 23g/L, more than 50\% attention weight is given, which means that the albumin level becomes the most critical indicator for one-year mortality outcome. 

On the other hand, between the range of 32-57 g/L, a high albumin value also causes high importance weight and indicates a significant improvement in the patient's health ($y<0.5$).
When the albumin level is higher than 40 g/L, it often occupies about 50\% to even 100\% of the feature importance weight, which means the model can predict the high survival expectation of patients using just this feature.
As a result, \mname recommends raising the albumin level to above 32 g/L as much as possible for most PD patients. 

The traditional clinical reference range of albumin for outpatient clinics is 40-55g/L, which is highly consistent with the recommended range given by \mname.
This finding is also consistent with the recent study that evaluated the association between serum albumin trajectories and mortality in PD patients using the joint modeling approach, showing that changes (increases and decreases) in serum albumin over time were strongly and significantly associated with mortality after adjustment for the risk factor \cite{basol2021effect}. 

\textbf{2) Diastolic Blood Pressure (DBP)} Fig.~\ref{fig:feature-importance-variation}-(b). 
DBP is another critical feature in the evaluation of patient health status. 
Similar to albumin, both high and low levels of DBP will affect the model's attention.
The importance weight of the DBP level varies in a V-shaped curve with 70 mmHg as a turning point. 
In the 40-70 mmHg range, the model pays higher attention to DBP level as it gets lower and predicts a poor prognosis.
When the DBP level is below 40 mmHg, it takes more than 30\% of the model attention weights.    
Most patients whose DBP level is below 60 mmHg are more likely to have a high health risk, marked as red dots in the figure. 

On the other hand, in the range of 70-120 mmHg, the model pays higher attention as the DBP level gets higher and predicts better prognosis outcomes.
When DBP is above 85 mmHg, it also occupies about 20\% of model attention weights, and patients are predicted in a low-risk condition for most cases marked as green dots in the figure. 
As a result, \mname recommends increasing the DBP level to above 70 mmHg, while greater DBP indicates lower risk.

This is consistent with recent studies on dialysis patients.
Higher DBP was associated with decreased early mortality in the first year after the start of renal replacement therapy (RRT) \cite{udayaraj2009blood}.
All-cause mortality risk was minimal at 77 mmHg for DBP in 7,335 Chinese peritoneal dialysis patients \cite{xie2020associations}.
DBP values less than 70 mmHg may be related to the increased mortality risk in both nondiabetic patients with chronic kidney disease patients and hemodialysis patients \cite{hannedouche2016multiphasic, robinson2012blood, navaneethan2017blood,xie2020associations}.
The traditional normal reference range of DBP for outpatient clinics is 60-80 mmHg, while maintaining DBP at a relatively higher level is conducive to improving the survival of PD patients. 
Further research about DBP for PD patients is needed.


\textbf{3) Systolic Blood Pressure (SBP)} Fig.~\ref{fig:feature-importance-variation}-(c).
Unlike the features discussed above, \mname believes that SBP is a typical feature whose importance weights vary in an L-shaped fold line with 130mmHg as a turning point, meaning that the importance weights decrease as the value increases.
For the SBP level below 60 mmHg, \mname gives more than 50\% attention, and in most cases, patients are likely to be predicted with poor outcomes presented as red dots in the figure.
For SBP level over 130 mmHg, \mname pays nearly no attention to SBP ($\alpha < 1\%$), which means that SBP does not affect the health status representation learning.
As a result, \mname recommends maintaining the SBP level at 130 mmHg or slightly higher for most PD patients. 
A further improvement over 130 mmHg does not significantly help reduce mortality risk.

This is consistent with clinic experience and most of the recent studies. 
Lower blood pressure was a surrogate marker for severe comorbid conditions (e.g., heart failure or ischaemic heart disease), chronic inflammation and malnutrition, and hence can lead to worse outcomes by limiting blood flow to vital organs~\cite{xie2020associations, afshinnia2016reverse}. 
The traditional reference range of SBP for outpatient clinics is 100-120 mmHg.
Although accepted definitions of hypertension and BP treatment targets in the dialysis population have not been determined and definitive recommendations regarding BP treatment targets in dialysis patients have not been made, it is clear that hypotension should be avoided \cite{flythe2020blood}. 

\textbf{4) Creatinine (Cr)} Fig.~\ref{fig:feature-importance-variation}-(e).
The importance variation curve of serum creatinine is also L-shaped, similar to SBP.
For Cr level in the 160-900 umol/L, the lower it is, the more attention is paid by \mname.
When the Cr level drops below 400 umol/L, the model provides more than 15\% of attention weights, and the patients are likely to face a poor prognosis ($y>0.5$).
For Scr level in the range of 900-1750 umol/L, it often only occupies 5\% of attention weights, and patients in this range generally have a lower mortality risk ($y<0.5$).
As a result, \mname recommends maintaining the Cr level at least 900 umol/L or slightly higher for most PD patients.

This is consistent with the finding of a previous study that a low Cr level (707-815 umol/L as reference) as a proxy of low muscle mass, nutritional status, and protein-energy wasting (PEW) may be associated with adverse outcomes in PD patients \cite{park2013serum, avram1995markers}. 
In contrast, a high Cr level is associated with a relatively lower mortality risk \cite{park2013serum}. 
Cr level should be maintained at a certain level.
The traditional reference range of Cr for normal outpatient clinics is 62-115 umol/L, which is unsuitable for PD patients.
Note that \mname provides a rough recommendation for most PD patients in this dataset. 
We will specify this finding for different cohorts (e.g., different gender) in future work.

\textbf{5) Hemoglobin (Hb)} Fig.~\ref{fig:feature-importance-variation}-(l).
The curve of the importance variation of hemoglobin is L-shaped. 
The model pays more attention to the hemoglobin level at 44-114 g/L as the hemoglobin level decreases. 
Hb occupies about 20\%-60\% of the model attention weights when the Hb level is below 100 g/L.
Patients in this range are more likely to suffer a high mortality risk. 
The model pays almost no attention to the hemoglobin level above 114 g/L.
As a result, \mname recommends keeping Hb level at least 114 g / L, but  further increment in Hb may not bring many benefits.

As indicated by a previous study, hemoglobin level lower than 100 g/L was significantly associated with a higher risk for all-cause and cardiovascular deaths\cite{kuo2018association}. 
Lower hemoglobin is also associated with a higher mortality risk in ESA-treated PD patients \cite{molnar2011association}.
Current anemia management guidelines also suggest ESAs not be used to maintain a Hb concentration above 110 g/L in adults~\cite{kliger2013kdoqi}.

More analyses about the features' importance are listed in Appendix section \ref{sec:add_featureimportamnce}.

\begin{figure}[]
  \centering
  \includegraphics[width=1.00\columnwidth]{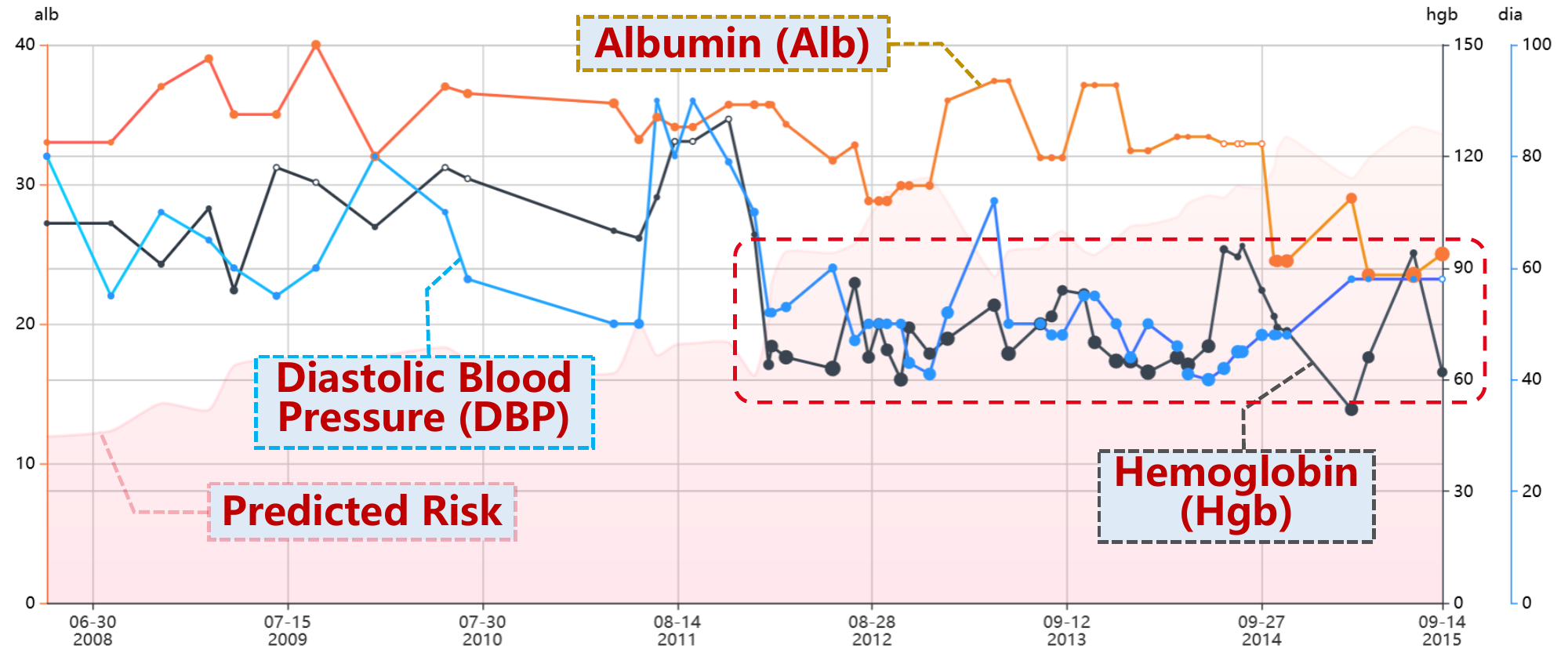}
  \caption{\textbf{Case Study I: Patient Died of Multiple Organ Failure.}
  Mortality risk prediction results and interpretability analysis are deployed on a health trajectory interactive visualization system.
  $x$-axis denotes the visiting date. 
  $y$-axis denotes the predicted \textcolor{pink}{mortality risk} (visualized as \textcolor{pink}{pink} translucent curve) and feature values.
  \mname provides the features' importance weights as interpretability at each visit, symbolized as the size of each data point (\textbf{·}) on the line chart and also listed in the suspension window (see Fig.~\ref{fig:overview}-(4) or check the link below).
 \mname pays most attention to \textcolor{orange}{Albumin}, \textbf{\textcolor{black}{Hemoglobin}} and \textcolor{blue}{Diastolic Blood Pressure} for this case patient, and the patient died of multiple organ failure, unfortunately. 
 The health trajectory interactive visualization system is publicly deployed at \url{http://47.93.42.104/A8}, which is available in English and Simplified Chinese.}
  \label{fig:useofmodel}
\end{figure}

\begin{figure}[]
  \centering
  \includegraphics[width=1\columnwidth]{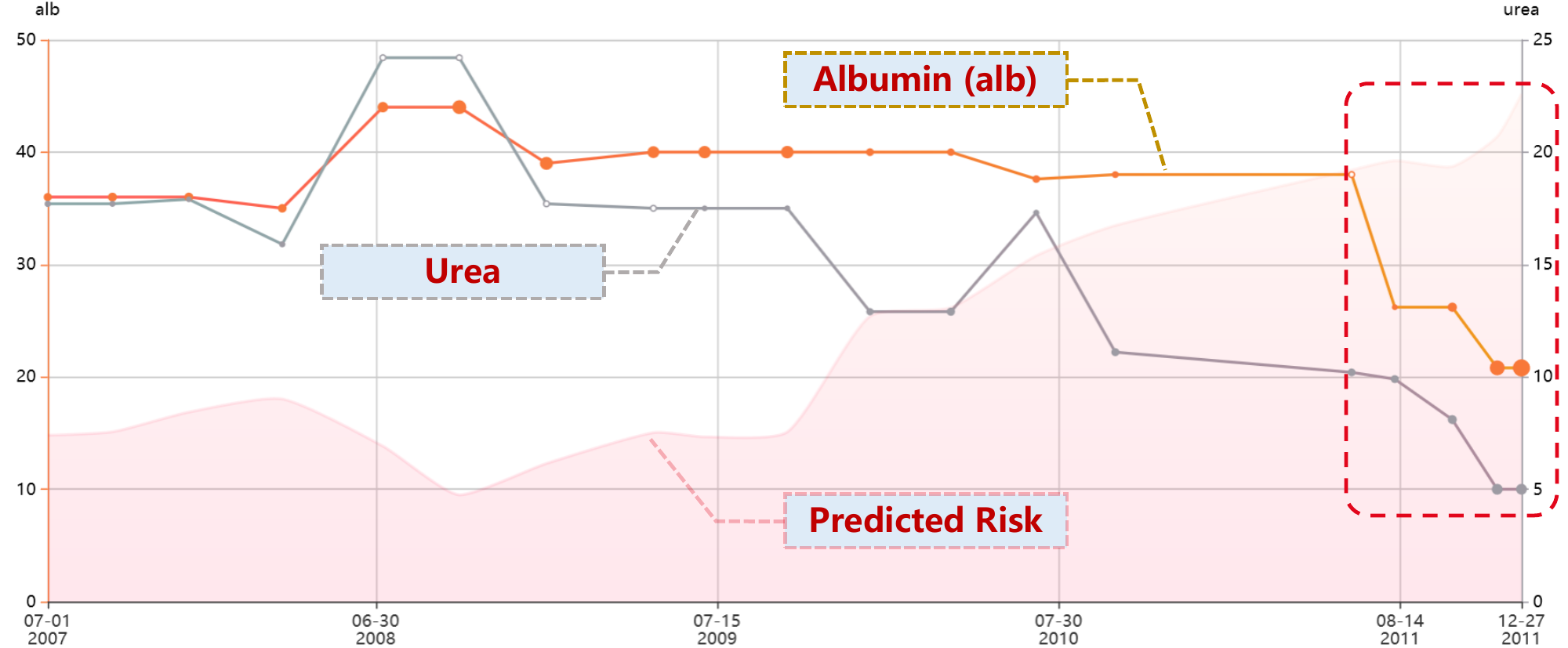} 
  \caption{\textbf{Case Study II: Patient Died of Digestive System Diseases.}
  \mname pays most attention to \textcolor{orange}{Albumin} and \textcolor{gray}{Urea} for this patient.
 The patient died of digestive system diseases, unfortunately. 
 The health trajectory is visualized at \url{http://47.93.42.104/A2}.}
  \label{fig:265}
\end{figure}

\subsection{Case Studies with Health Trajectory Interactive Visualization System}

To intuitively show the prediction process and verify the reasonability of \mname when applied to clinical practice, 
we develop an AI-Doctor online system with an interactive interface to visualize the patient's health trajectory with the importance weights of features at each timestep.
This system makes the prediction results of deep learning models more accessible to clinicians and helps physicians make individualized clinical decisions.
We draw a two-dimensional line chart to show the changes in the patient's biomarkers. 
The $x$-axis is the visit timeline and the $y$-axis is the value of biomarkers.
At each time step, we plot the predicted risk curve $\hat{y}_i$ (values from 0-100\%).
The attention weights of different biomarkers at each time step are also visualized, symbolized as the size of each data point on the line chart. 
The larger the point, the higher the attention weight.

In the following subsections, we analyze two patient cases using our system.
The clinical visit dates were reset to start in 1000 (year) on the online visualization system to protect privacy.
For ease of understanding, the dates in this manuscript are presented as original, which will also be modified when published.

\noindent \textbf{Case I: Patient Died of Multiple Organ Failure (Fig.~\ref{fig:useofmodel})}

The first case was a 66-year-old male. 
He was diagnosed with diabetic nephropathy and initiated PD therapy on December 10, 2007.
He died on September 24, 2015, due to prostate cancer and multiple organ failure. 
Fig.~\ref{fig:useofmodel} shows the patient's risk prediction and historical visit information. 

During the period that the red dotted box covers in the figure, \mname kept predicting a high risk three years before the adverse outcome. 
\mname mainly focused on albumin, diastolic blood pressure (DBP) and hemoglobin (Hb) due to their abnormal values and declining patterns. 
It is evident that the value of Hb and DBP decreased sharply at the beginning of 2012, which decreased by 69 g/L (from 130 g/L to 61 g/L) and 27 mmHg (from 79 mmHg to 52 mmHg), respectively.
\mname sensed the change rapidly and started to pay attention to them. 
There were 31.0\% of attention given to Hb, and 19.8\% given to DBP.
We can also find a sudden drop of albumin from 32.9 mmol/L to 24.5 mmol/L in October 2014, and the albumin level remained at a low level during the last several visits since then, which kept drawing 30\%-40\% of attention weights of our model. 

According to the records, this patient had a series of comorbidities since 2012, including unstable angina pectoris, peripheral arterial disease (PAD), prostate cancer, anemia, diabetic foot, and inflammatory bowel disease, which were closely related to the abnormal biomarkers warned by \mname.
Specifically, the decline of DBP indicated worsening arterial stiffness, which may be associated with severe atherosclerosis in these patients such as coronary heart disease, PAD and diabetic foot occurred in this patient. 
The abnormal hemoglobin level indicated deleterious anemia and could be associated with gastrointestinal bleeding, severe infection, malnutrition, prostate cancer, diabetic foot and inflammatory bowel disease \cite{de2009association,cheng2005strong,tsirpanlis2005serum}.
With the help of \mname, physicians may be early reminded to do a further examination to confirm and treat the upper-mentioned situations accordingly.

\noindent \textbf{Case II: Patient Died of Digestive System Diseases (Fig.~\ref{fig:265})}

The second case was a 68-year-old female. 
As shown in Fig.~\ref{fig:265}, she was diagnosed with ischemic kidney disease and initiated PD therapy on June 10, 2005. 
She died on February 7, 2012, due to gastrointestinal disease. 

Since October 1, 2010, the risk score generated by \mname has continuously increased. 
The attention varied but mainly focused on serum chloride, sodium and urea level, which were indicators of insufficient intake or gastrointestinal loss (The details are visualized at \url{http://47.93.42.104/A2}). 
On November 30, 2011, 53.9\% AI attention was assigned to serum albumin (albumin level decreased from 38g/L to 20.8g/L). 
Finally, in the last visit on December 27, the risk score for this patient was 90.3, and 66.2\% attention was given to albumin level. 
This patient went to the emergency room on January 27, 2012, and died on February 7, 2012, due to an unknown GI disease.
\mname captured the most important clinical features related to patient death and generated a high-risk score timely.

\section{Discussion}

\subsection{Implications}

The whole procedure of the PD treatment needs a dynamic prediction of patient mortality risk to help patients prevent adverse outcomes, based on the medical records collected along with the visits.
Individual-level dynamic mortality prediction for long-term PD has not yet been substantially studied.
Besides, deep models, which can capture complex longitudinal progressions, are often black boxes and fail to provide human-understandable interpretation.
Thus, medical professionals lack trustworthiness in deep models.

In this work, we develop a deep-learning-based generalizable model capable of learning massive EMR data and exploring personal characteristics to perform clinical predictions.
\mname captures the clinical features that strongly indicate the health status of patients in various conditions.
It personally builds health status embedding and provides reasonable fine-grained interpretability in terms of feature importance at each follow-up visit.

In the experiment based on a real-world clinical dataset, 656 incident peritoneal dialysis patients were enrolled at Peking University Third Hospital. 
\mname is used to predict one-year mortality at each follow-up visit. 
We compare the performance of \mname against existing related state-of-the-art (SOTA) clinical predictive models. 
The experiment results show that \mname outperforms the published baseline approaches with 11.8\% relative improvement on AUPRC and powerful interpretability.

To facilitate personalized clinical service and verify the reasonability of the model, we develop an AI-Doctor interaction system to reveal the patient's health trajectory and the corresponding vital biomarkers while performing a prognosis. 
After the trial of our system, experienced nephrology department physicians suggest that \mname can offer opportunities to identify patients with potential mortality risks within a time window that enables early individualized treatment and outcome improvement.
The medical knowledge learned by \mname has been positively confirmed by human medical experts and related medical literature.


\subsection{Key Findings and Clinical Recommendations Restatements}

\subsubsection{Important Features Summary}

Some of the key findings generated by \mname are summarized below.
For more details about the medical findings, please check Fig~\ref{fig:feature-importance-variation} and Table~\ref{tab:turningpoint}.

\begin{itemize}[leftmargin=*]

\item \textbf{Albumin} is the most indicative feature for the prediction of 1-year mortality in patients with PD, especially for gastrointestinal (GI) disease, peripheral vascular disease (PVD) and alive patients. 
The feature importance weight of albumin presents as a V-shaped curve along with the albumin level.
A higher albumin level brings better survival expectations. 
\mname recommends raising the albumin level to above 32 g/L as much as possible for most PD patients. 

\item \textbf{Diastolic Blood Pressure (DBP)} is the second important feature. 
It is indicative especially for cachexia, cancer, cerebrovascular disease (CVE) and alive patients.
\textbf{Systolic Blood Pressure (SBP)} is indicative for cancer and peritoneal dialysis-associated peritonitis (PDAP).
The importance weight of DBP and SBP presents as V-shaped and L-shaped curves, respectively.
\mname recommends raising the DBP level to above 70 mmHg for most PD patients.
\mname recommends maintaining the SBP level at least 130 mmHg. But further increment of SBP will not bring many benefits. 

\item \textbf{Chlorine (Cl)} is indicative for cachexia and infection patients.
The importance weight of Cl presents as a V-shaped curve.
A Higher Cl level brings better survival expectations.
\mname recommends raising the Cl level to above 96 mmol/L for most PD patients.

\item \textbf{Creatinine (Cr)} is indicative for GI disease and cardiovascular disease (CVD) patients. 
\mname recommends raising the Cr level to above 900 umol/L, 
which is a rough recommendation for most PD patients in this dataset. 
We will specify this finding for different cohorts (e.g., different gender) in future work.
\textbf{Urea} is indicative for PDAP and PVD patients.
\mname recommends raising urea level to above 20 mmol/L for most PD patients.

\item \textbf{Phosphorus (P)} is indicative for PDAP, cancer, CVD and CVE patients.
The importance weight of P presents as an L-shaped curve.
\mname recommends raising P level to above 1.5 mmol/L for most PD patients. Further increments will not bring benefits.

\item \textbf{Hemoglobin (Hb)} is indicative for GI disease patients.
The importance weight of Hb presents as an L-shaped curve.
\mname recommends raising Hb level above 114 g/L for most PD patients. 
Further increments will not bring many benefits.

\end{itemize}

\subsubsection{Prediction Performance of Mortality Causes}

As shown in Fig.~\ref{tab:statistics_COD} in Appendix Section \ref{sec:cod}, experiment results indicate that \mname can effectively predict the most common adverse outcomes of PD patients (e.g., cachexia, peripheral vascular disease, infection, cancer).
However, cerebrovascular disease (CVE) and cardiovascular disease (CVD) are the most challenging mortality causes to predict. 
Particularly CVE \cite{wiki:cerebro} patients often acutely suffer from sudden death without apparent signs.
This suggests more frequent clinical follow-ups and more clinical tests included as features (e.g., ECG examination) to perform early warning for CVE.

\subsection{Limitations and Future Works}

\subsubsection{Introduce Multi-Center EMR to Increase Data Amount}

A major limitation of this study is the single-center design, which makes the data amount of the research scarce. 
The limitation also results in a relatively small sample of positive cases. 
However, the analyzed data of 656 PD patients with 13,091 visits cover a long-term longitudinal trajectory of PD patients.
There are about 20 visits recorded for each patient, with an average visit interval of 2.7 months and an average follow-up time of 4 years.
To the best of our knowledge, this is rarely seen in the existing medical literature.
Besides, we also introduce a hemodialysis EMR dataset as an additional experiment dataset to test the prediction performance.
In future work, we will extend \mname to multi-center healthcare systems and conduct a prospective multi-center controlled experiment to validate the framework in other clinical scenarios.

\subsubsection{Build Practical Application Simulation}

Although it is widely believed that accurate predictions can be used to improve care~\cite{bates2014big},
this is not a foregone conclusion and prospective trials are needed to demonstrate this~\cite{krumholz2016data,grumbach2014transforming}.
We will conduct a blinded application-grounded evaluation by inviting dozens of experienced medical practitioners (with 7-20 years of clinical practicing time) from nephrology departments of different hospitals to test the practical effectiveness and physicians' recognition degrees.
The prototype version of the trial system with online questionnaires has already been developed at \url{http://47.93.42.104/table/questionnaire/a1}. 
The AI-Doctor interaction system is available in English and Simplified Chinese. The questionnaire page is currently only available in Simplified Chinese.

\subsubsection{Incorporate More Clinical Features to Depict Health Status}

During the data collection process for this study, we collected many medical features of patients, most of which were discarded due to high missing rates. 
Our model had access to only the auto-filtered 16 longitudinal medical features and 4 demographic features for each patient.
The novelty of this research does not only lie in incremental model performance improvements. 
This predictive performance was achieved without hand-selection or hand-made variables deemed important by a medical expert.
\mname can achieve satisfactory prediction results and discover medical findings, proving the model's validity and practicability.
In future releases, we expect to incorporate more medical features, such as medication records, dialysis adequacy records, complication records and health data collected at home.

\subsubsection{Provide Recommendation for Diverse Patient Cohorts}

In order to obtain relatively stable and reasonable conclusions, the clinical recommendations in this paper are roughly generated by \mname for most PD patients. 
As more data is collected in the next release, we will provide refined recommendations for diverse patient cohorts (e.g., different genders, ages).


\subsubsection{Embed Feature Sequences Properly}

As shown in Fig.~\ref{fig:feature-importance-variation}, no significantly meaningful importance pattern was discovered for white blood cell count (WBC).
This may be because WBC is not a crucial feature in mortality prediction, or WBC is such a special clinical feature that \mname does not know how to use it to embed the health representation.
This reminds us to design proper embedding network modules (e.g., convolutional layers) to utilize different feature sequences effectively.
Besides, considering that the proportion of immune cells may indicate the health status as a humanly constructed advanced feature, we will introduce more related clinical features about the immune. 

\appendices

\section{Materials and Method Details}

\subsection{Dataset}
\label{sec:dataset}

\subsubsection{Statistics of Age and Visit Frequency}

\begin{figure}[]
  \centering
  \includegraphics[width=0.95\columnwidth]{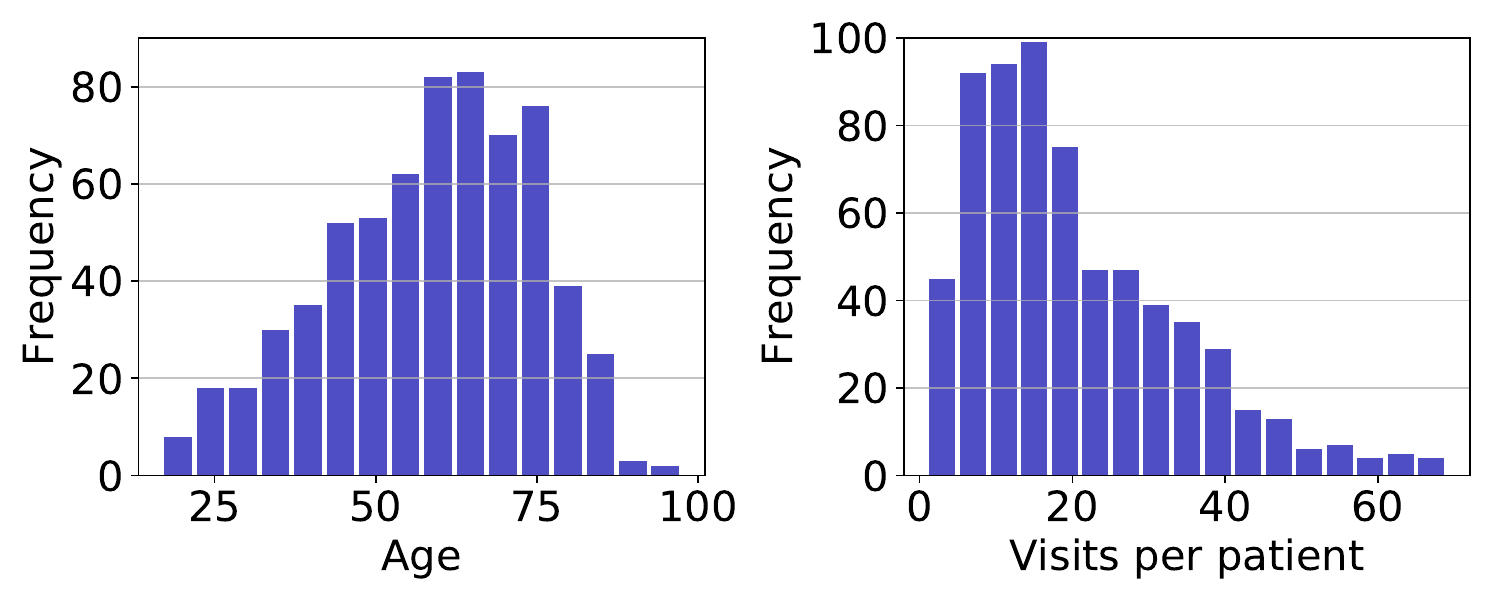}
  \caption{\textbf{Distribution of Age and Visit Frequency in PD Dataset.} }
  \label{fig:data_distrib}
\end{figure}

This work includes 13,091 visits of 656 end-stage renal disease peritoneal dialysis patients from Peking University Thirds Hospital.
Fig.~\ref{fig:data_distrib} shows the distribution of age and visit frequency. 
The average age of patients at the first clinical visit is 58.55 years old, with a standard deviation (Std) of 15.81 years.
The average visiting frequency of patients at the end of the clinical follow-up was 19.95, with an Std of 13.53.

\subsubsection{Ethic and Privacy Issues}

This study was approved by the Medical Scientific Research Ethical Committee of Peking University Third Hospital. 
The approval document is attached to the submitted files.
Patients' private information was anonymized during the analysis. 
Patient names were replaced with unique patient IDs (e.g., A1, A2).
Contact information, including phone numbers and addresses, was deleted.
Clinical visiting dates were reset to start in 1000 (year) on the online visualization system, and the date of birth was also reset with the corresponding offset.
For ease of understanding, the health trajectory visualization figures in this paper are presented in original dates, which will also be rest when published.

\subsection{Problem Formulation}
\label{sec:problem_formulation}

We formulate the model inputs and prediction tasks as follows:
\begin{definition}[\textbf{Patient Records}]
A patient's visit records can be represented as a matrix $\mathbf{R}\in \mathbb{R}^{N\times T}$, where $N$ denotes the number of medical features in visit records and $T$ denotes the number of visits within the observation window.
We use vector $\mathbf{r}_{n,t}$ to denote the $t$-th visit of the $n$-th feature. 
The baseline information is denoted as vector $\mathbf{r}_0$.
\end{definition}

\begin{figure}[]
  \centering
  \includegraphics[width=0.95\columnwidth]{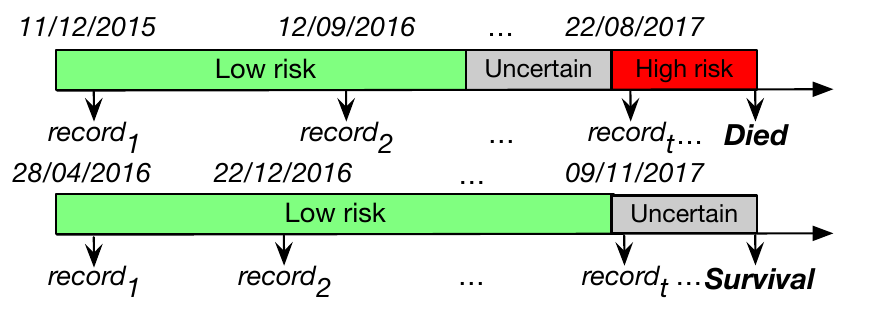}
  \caption{\textbf{Label Assignment.} The prediction task is defined as a 1-year mortality prediction at each clinical visit. Clinical visits within 1 year before death are labeled as \textit{high risk} ($y=1$). Visits recorded 2 years before death are labeled as \textit{low risk} ($y=1$). Visits recorded between 1 and 2 years before death are labeled as \textit{uncertain} status and will not be included in the training process.}
  \label{fig:status}
\end{figure}

\begin{problem}[\textbf{One-year Mortality Prediction}]

Given a patient's visit records $\mathbf{R}$ and baseline information $\mathbf{r}_0$, our objective is to predict the mortality risk $\widehat{\mathbf{y}}_t$ in the next year for the patient at each visit. 
This is formulated as a binary classification task as $\mathbf{y} \in \left \{ 0, 1 \right \}$. 

Considering the uncertainty of the health status variation in the observation window, we design a particular labeling strategy to make the training labels as close to the ground truth as possible. 
As shown in Fig.~\ref{fig:status}, for patients with positive labels (i.e., mortality) at the end of clinical follow-up, we consider all visits within one year before the mortality date as high-risk visits ($y = 1$). 
For patients with negative labels at the end, we consider all visits within one year before the last visit as \textit{uncertain}, since we do not know whether the patient will have an adverse outcome in the next year ($y = uncertain$).
The calculation of loss function and performance metrics will not include these visits.
Other visits are all labeled as low-risk ($y = 0$).
\end{problem}

\begin{problem}[\textbf{Model Interpretability}]
For a given patient, the model will output an importance score matrix $\mathbf{\alpha}$, where the value $\alpha_{n,t}$ in the $n$-th row and $t$-th column denotes the importance score for feature $n$ at visit $t$. 
This importance score represents how much the feature contributes to the prediction. 
\end{problem}




\subsection{Model Detail}
\label{sec:model_detail}



We propose a general healthcare predictive model, which can adaptively depict patients' health status in diverse conditions and provide reasonable interpretability. 
The model explicitly captures the interdependencies among time series of dynamic features and static baseline information to learn the personal health context of patients in a global view.
As shown in Fig.~\ref{fig:overview}-(2), \mname comprises the following sub-modules:

\begin{itemize}
    \item The multi-channel feature extraction module is developed to learn the representation of each dynamic feature separately.
    
    \item The adaptive feature importance recalibration module enhances the key factors explicitly by squeeze-and-excitation block to perform the individualized clinical prediction.
    
\end{itemize}

\subsubsection{Multi-Channel Feature Extraction}

The patient's health status is depicted by dynamic feature sequences and static baseline features.
To embed such heterogeneous information and build the final health representation, as well as make the dynamic feature's importance assignment process more intuitive, we develop a multi-channel feature extraction module.

Considering that the logical time order of the clinical sequence matters in the medical domain, \mname embeds the time series of each dynamic feature separately by multi-channel bidirectional GRUs.
The GRUs at the bottom level will refine the representation of each feature and form a sequence of $N+1$ feature vectors ($N$ for dynamic record features $\mathbf{R}$ and $1$ for the baseline information $\mathbf{r}_0$). 
Specifically, \mname embeds the time series of each feature separately by multi-channel bidirectional GRU:
\begin{equation}
(\overrightarrow{\mathbf{f}_{n,1}}, ..., \overrightarrow{\mathbf{f}_{n,T}}), (\overleftarrow{\mathbf{f}_{n,1}}, ..., \overleftarrow{\mathbf{f}_{n,T}}) = \text{bi-GRU}_n(\mathbf{r}_{n, 1}, ..., \mathbf{r}_{n, T}), 
\end{equation}
where the time series of feature $n$ is denoted as 
 $\mathbf{r}_{n,:} = (\mathbf{r}_{n,1}$, ...,$ \mathbf{r}_{n,T} )$ 
$\in {R}^{T}$.
We derive each feature representation as the sum of the two last hidden embeddings:
\begin{equation}
\mathbf{f}_{n} = \overrightarrow{\mathbf{f}_{n,T}} + \overleftarrow{\mathbf{f}_{n,1}}.
\end{equation}
Furthermore, the demographic baseline data ($\mathbf{r}_0$) is embedded as:
\begin{equation}
\mathbf{f}_{0} = W^{emb}_{0} \cdot \mathbf{r}_0,
\end{equation}
where $W^{emb}_{0}$ is the embedding matrix. 
From here on, we ignore the bias term for ease of notation.
Thus, all the patient data can be represented by a matrix $\mathbf{F}$ (i.e., a sequence of vectors, where each vector represents one feature of the patient over time):
$\mathbf{F}=(\mathbf{f}_{0},\mathbf{f}_{1},\cdots,\mathbf{f}_{N})$.

\subsubsection{Adaptive Feature Importance Recalibration}

We develop an adaptive feature importance recalibration module to provide the feature's importance weight when performing mortality prediction at each clinical visit.
This attention-based module is inspired by the $SEblock$ in the research area of computer vision \cite{hu2018squeeze}, and is trained to explicitly model the dependencies between clinical features. 
It can selectively give more weight to the representative and predictive features but suppresses unimportant ones.

First, to select the most informative features, we should provide the model with a global view of patients' status at the current visit. 
For a particular patient, \mname squeezes all the feature embedding via a mean pooling operation to get an integrated but straightforward picture of the health status. 
It can be regarded as an abstract of the patient's all historical status. 
The importance of different temporal patterns will be calculated based on this abstract.
Second, we selectively give more weight to the predictive features but suppress unimportant ones, which contribute little to the prediction target.
The selectively enhanced predictive features can be treated as a precursor of health risk for the given patient.

Concretely, the $Query$ is obtained by $\mathbf{f}_{sqz}$ from embedded health information $\mathbf{F}$, including dynamic features and baseline features.
The $Key$s are formed by embedded dynamic features $\mathbf{f}_{1},\cdots,\mathbf{f}_{N}$ as:

\begin{equation}
    \mathbf{f}_{sqz} = \operatorname{Mean\_pooling}(\mathbf{f}_{0},\mathbf{f}_{1},\cdots,\mathbf{f}_{N}),
\end{equation}
\begin{equation}
    \mathbf{q}^{fin}_{sqz} = W^{fin}_{sqz} \cdot \mathbf{f}_{sqz},
\end{equation}
\begin{equation}
    \mathbf{k}^{fin}_{n} = W^{fin}_{n} \cdot \mathbf{f}_{n}, \,\,\,\, (n= 1, ..., N),
\end{equation}
where $W^{fin}_{sqz}$ and $W^{fin}_{n}$ are the projection matrix respectively. 
Then the attention weights are calculated as follows:

\begin{equation}
 \mathbf{\alpha}^{fin}_{1}, ..., \mathbf{\alpha}^{fin}_{N} = \delta( \mathbf{\zeta}^{fin}_{1}, ..., \mathbf{\zeta}^{fin}_{N}),
\end{equation}
\begin{equation}
\mathbf{\zeta}^{fin}_{n} =  \mathbf{q}^{fin}_{sqz} \cdot \mathbf{k}^{fin}_{n}  \,\,\,\, (n= 1, ..., N),
\end{equation}
where $\delta$ denotes the activation function (i.e., \textit{softmax} or \textit{sparsemax}).
\textit{Sparsemax} activation will make the model interpretability more prominent by allowing the most critical features to dominate the final embedding. 
Using \textit{sparsemax} activation will slightly weaken the performance since it suppresses too much information from features that are not most important. 
In this work, we employ \textit{softmax} function to obtain the prediction results, and employ \textit{sparsemax} function to extract medical findings.
Finally, the health status representation $s$ and the prediction result $\widehat{\mathbf{y}}$ can be obtained by:
\begin{equation}
\mathbf{s} = \operatorname{Concat}[ \sum_{n=1}^{N} \mathbf{\alpha}^{fin}_{n}\cdot \mathbf{f}^{*}_{n}, \, \mathbf{f}^{*}_{0}],
\end{equation}
\begin{equation}
\widehat{\mathbf{y}} = \operatorname{Sigmoid}(W^{final} \cdot \mathbf{s} ),
\end{equation}
where $W^{final}$ is the weight matrix. 

\section{Experiment Details}
\label{sec:Experiment_Details}

\subsection{Experiment Setup}

\subsubsection{Implementation Details}

The training is done in a machine equipped with CPU: Intel Xeon E5-2630, 256GB RAM, and GPU: Nvidia Titan V using Pytorch 1.1.0. 
We use Adam \cite{kingma2014adam} with the mini-batch of 256 patients, and the learning rate is set to 1e-3. 

\subsubsection{Baseline Approaches}

We implement several state-of-the-art (SOTA) representative models as comparative baseline approaches.
The hyper-parameters of models are fine-tuned by a grid-searching strategy. 

\begin{itemize}[leftmargin=*]

\item \textbf{GRU}: Gated Recurrent Unit neural network embeds the time series to perform the target prediction.
It is a widely applied variant of the Recurrent Neural Network (RNN), which improves the capability to maintain historical memories and reduces parameters in the update and reset gates.
GRU has been used to predict several severe complications (mortality, renal failure with a need for renal replacement therapy, and postoperative bleeding leading to operative revision) in post-cardiosurgical care in real-time (Lancet Respiratory Medicine, 2018) \cite{meyer2018machine}.

\item \textbf{Transformer$_e$} is the encoder of the Transformer~\cite{vaswani2017attention}, which comprises the positional encoding module and the self-attention module.
Transformer has been used to perform the mortality risk analysis for liver transplant recipients~\cite{nitski2021long}.

\item \textbf{MT-RHN}: Multi-Task Deep Recurrent Highway Network, which embeds the historical data and current step with a deep residual embedding component, and employs a multi-task learning strategy to enhance the performance.
MT-RHN has been used to perform the continuous risk prediction of future acute kidney injury deterioration occurring in the next 48 hours (Nature, 2019) \cite{tomavsev2019clinically}.

\item \textbf{LSTM}: Long Short Term Memory network is a variant of the Recurrent Neural Network (RNN), capable of learning long-term dependencies.
LSTM has been used to perform the 90-day all-cause mortality in the intensive care unit (ICU), based on the concatenated static features and dynamic features (Lancet Digital Health, 2020) \cite{thorsen2020dynamic}.

\item \textbf{biLSTM-FC}: Bidirectional LSTM with Fully Connected layers.
Bidirectional LSTM (biLSTM) is a sequence processing model that consists of two LSTMs: one taking the input in a forward direction and the other in a backward direction.
Sung et al. have used biLSTM-FC to perform the clinical event prediction (death, sepsis, and acute kidney injury), where biLSTM and fully connected layers are employed to embed the dynamic features and static features correspondingly (JMIR, 2021) \cite{sung2021event}.

\item \textbf{XGBoost}: a recursive tree-based supervised machine learning classifier.
XGBoost has been used to predict the mortality for COVID-19 infected patients (Nature Machine Intelligence, 2020) \cite{yan2020interpretable}.

\item \textbf{DT}: Decision Tree, a non-parametric supervised learning algorithm with a hierarchical tree structure.
DT has been used to perform mortality prediction for peritoneal dialysis patients (Nature Scientific Reports, 2020) \cite{noh2020prediction}.


\item \textbf{LightGBM}: a gradient boosting ensemble framework that uses tree-based learning algorithms.
LightGBM has been used to perform the early prediction of circulatory failure in the intensive care unit (Nature Medicine, 2020) \cite{hyland2020early}.

\item \textbf{LR}: Logistic Regression model predicts a dependent data variable by analyzing the relationship between one or more existing independent variables.
LR has been used to predict the early risk of chronic kidney disease in patients with diabetes (Nature Medicine, 2019) \cite{ravizza2019predicting}.


\end{itemize}


\subsubsection{Evaluation Metrics}

We evaluate the models with a 10-fold cross-validation strategy and report the average performance, similar to \cite{ma2018health}.
We assess performance for the binary classification problem using the area under the receiver operating characteristic curve (AUROC) and the area under the precision-recall curve (AUPRC).
AUPRC is the most informative and the primary evaluation metric when dealing with a highly imbalanced and skewed dataset \cite{davis2006relationship,choi2018mime} like the real-world EMR data.

\subsection{Prediction Performance of \mname for Different Causes of Death on PD Dataset}
\label{sec:cod}

\begin{table*}
\centering
  \caption{\textbf{Statistics of patients with different mortality causes.} 
  This real-world dataset contains 656 peritoneal dialysis (PD) patients.
 261 (39.8\%) patients of them, unfortunately, died before the final follow-up. 
  There are nine different causes of death (COD) recorded.  
}
  \label{tab:statistics_COD}
  \begin{tabular}{lrc}
     \hline
     Causes of Death &  \# Patients (\%) & AUROC \\
     \hline
    Cerebrovascular Disease (CVE) & 74 (28.3\%) & 0.55 \\
    Cardiovascular Disease (CVD) & 21 (8.0\%) & 0.71  \\
    Gastrointestinal Disease (GI Disease) & 17 (6.5\%) & 0.73  \\
    Peritoneal Dialysis Associated Peritonitis (PDAP) & 21 (8.0\%) & 0.74  \\
    Cancer & 23 (8.8\%)  & 0.76 \\
    Other & 50 (19.1\%) & 0.80  \\
    Infection & 33 (12.6\%)  & 0.82 \\
    Peripheral Vascular Disease (PVD) & 13 (4.9\%) & 0.82  \\
    Cachexia & 9 (3.4\%)  & 0.88 \\
     \hline
    Mortality & 261 (100.0\%)  & - \\
    \hline
\end{tabular}
\end{table*}

There are nine different CODs recorded: Cerebrovascular Disease (CVE), Cardiovascular Disease (CVD), Peritoneal Dialysis Associated Peritonitis (PDAP, Peritonitis), Peripheral Vascular Disease (PVD), Infections, Gastrointestinal Disease (GI Disease), Cachexia, Cancer and Other causes.
We evaluate the performance for patients with different causes of death (COD).
Since our experiment was conducted via 10-fold cross-validation, we employ the model trained on each fold's training set to the corresponding testset to perform the prediction.

\begin{figure}[]
  \centering
  \includegraphics[width=0.8\columnwidth]{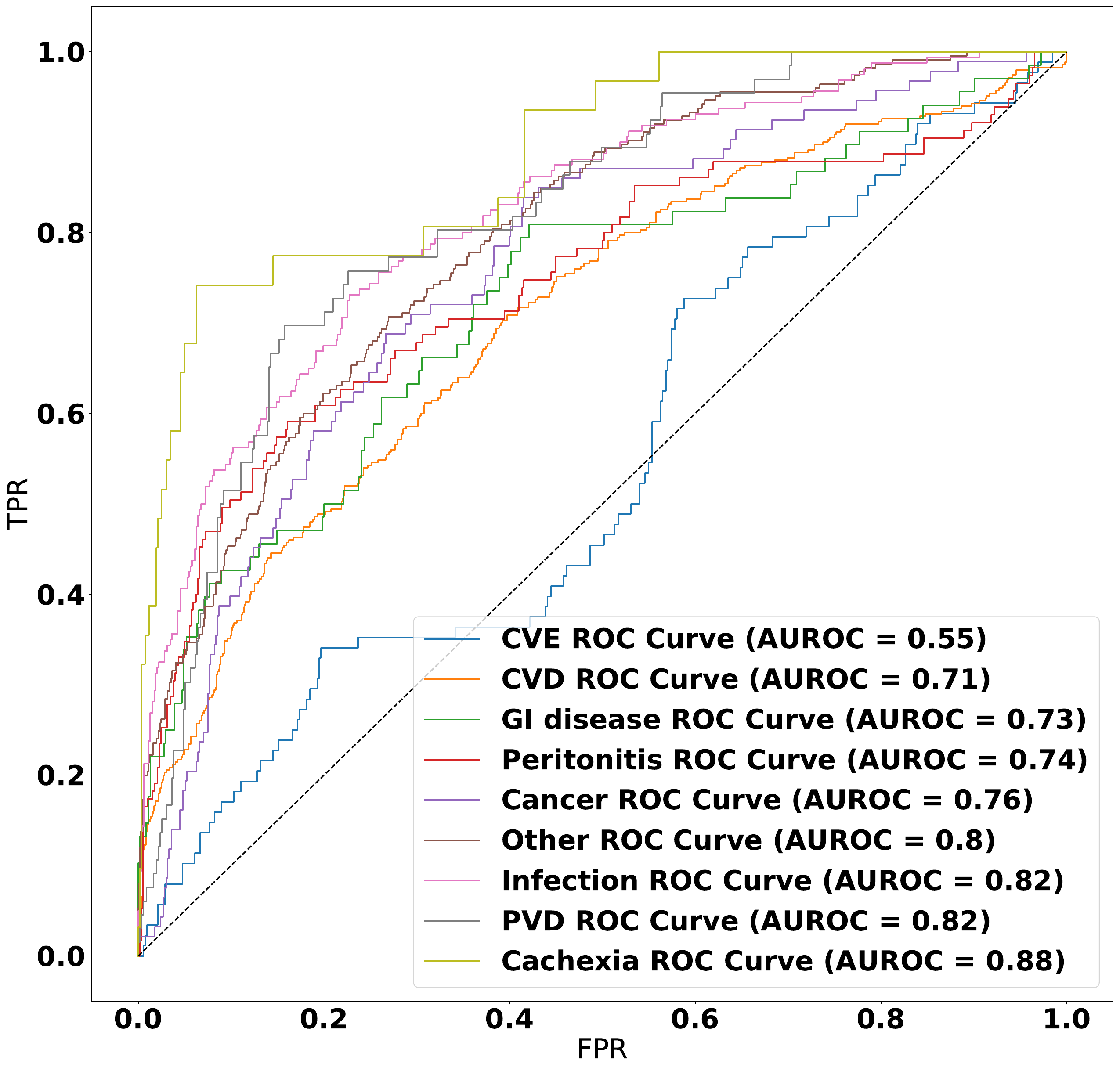}
  \caption{\textbf{Prediction ROC Results of Different Mortality Causes.}
Cerebrovascular Disease (CVE) and Cardiovascular Disease (CVD) are the most challenging mortality causes to predict. On the contrary, Infections, Peripheral Vascular Disease (PVD), and Cachexia risks are relatively easy to be early identified. }
  \label{fig:results-cod}
\end{figure}

The statistics of patients with different mortality causes are shown in Table~\ref{tab:statistics_COD}.
The receiver operating characteristic (ROC) curves for different COD patient subgroups are shown in Fig.~\ref{fig:results-cod}.
According to the prediction results, the risk of cachaxia (AUROC = 0.88), infection (AUROC = 0.82) and PVD (AUROC = 0.82) are easy to be identified. 
\mname provides accurate prediction results of these patients about a year before the outcome.
 
On the contrary, patients with CVE (AUROC = 0.55) and CVD (AUROC = 0.71) are the most difficult to predict by the model. 
These diseases often attack untimely and acutely without obvious signs \cite{wiki:cerebro}, compared to cachexia, infections, and PVD.
Patients with these health risk factors may have a higher threat and probability of sudden death in quite a short period of time, which is hard to take early warnings.

\subsection{Additional Experiment on External Dataset}
\label{sec:HDresult}

\mname is a generic framework proposed to model the patient's health status on multi-variate time series EMR data.
The analysis of the peritoneal dialysis dataset in this paper serves as a proof of concept.
To verify the generalizability of \mname, we train the model to perform the mortality prediction on an external hemodialysis ESRD dataset.
This is a real-world EMR collected at Peking University People's Hospital.
The statistics of the dataset are listed in Table~\ref{tab:hd-statistics}.
The results in Table~\ref{tab:performance-external} indicate that \mname also achieves better performance than the SOTA models.

\begin{table}
\centering
  \caption{\textbf{Statistics of Hemodialysis Patient EMR as an Additional Dataset.} The real-world dataset contains 1,363 hemodialysis dialysis (HD) patients with 4,789 clinical visits. 
  There are 12.55\% patients, unfortunately, who died before the final follow-up. 
  The average age of patients enrolled is 59 years old.  
}
  \label{tab:hd-statistics}
  \begin{tabular}{lccc}
     \hline
     & Total & Survival (\%) &  Mortality (\%) \\
     \hline
    \# Patients & 1363 & 1192 (87.45\%) & 171(12.55\%)\\
    \# Visits & 4789 & 4359 (91.02\%)& 430 (8.98\%)\\
    Avg. Age & 59.38 & 57.89 & 68.67 \\ 
    \hline
\end{tabular}
\end{table}

\begin{table*}[]
\small
 \centering
  \caption{\textbf{Feature Summary of Hemodialysis (HD)
 Dataset.} This dataset comprises 18 dynamic features recorded at each clinical visit and 4 static baseline features recorded at the first visit.}
  \label{tab:feature_summary_hd}
  \begin{tabular}{ccccccccccc}
    \hline
     & Abbrev. & Full Name & Unit & \multicolumn{3}{c}{Low Risk Visits ($y=0$)}   & \multicolumn{3}{c}{High Risk Visits ($y=1$)} & \% Missing  \\
    \hline
     & \multicolumn{2}{l}{Dynamic Features}   & & Mean & Std & Median & Mean & Std &Median &\\
    \hline
    & CO$_2$CP &  CO$_2$ Combining Power& mmol/L & 21.579 & 3.867 & 21.6 & 21.292 & 4.066 & 21.4 & 27\% \\
    & WBC &  White Blood Cell Count& x10$^9$/L & 6.216 & 1.981 & 5.950 & 6.254 & 2.391 & 5.85 & 6\% \\
    & Hb &  Hemoglobin &g/L & 107.9 & 17.283 & 109.5 & 103.6 & 18.66 & 105 & 6\% \\
    & Ca &  Calcium & mmol/L & 2.272 & 0.24 & 2.258 & 2.253 & 0.236 & 2.24 & 41\% \\
    & K &  Potassium & mmol/L  & 5.045 & 0.818 & 5 & 4.851 & 0.851 & 4.8 & 13\% \\
    & Na &  Sodium& mmol/L & 139.1 & 3.451 & 139 & 138.6 & 3.437 & 138.9 & 16\% \\
    & Cr &  Creatinine & umol/L & 933.4 & 296.7 & 917.8 & 784.6 & 306.0 & 740.3 & 17\% \\
    & P &  Phosphorus& mmol/L & 1.896 & 0.636 & 1.811 & 1.817 & 0.671 & 1.701 & 15\% \\
    & Albumin &  Albumin  &g/L & 39.49 & 4.212 & 39.9 & 36.94 & 4.838 & 37.5 & 27\% \\
     & Glucose &  Glucose& mmol/L & 7.131 & 3.677 & 6.11 & 7.846 & 4.361 & 6.68 & 38\% \\
    & pre-Weight & Pre-Dialysis Weight & kg & 62.35 & 12.24 & 61.5 & 60.57 & 11.91 & 60.01 & 61\%  \\
    & pst-Weight & Post-Dialysis Weight & kg & 59.94 & 11.96 & 59.06 & 58.23 & 11.58 & 57.23 & 62\%  \\
    & pre-SBP & Pre-Dialysis SBP & mmHg & 147.3 & 20.58 & 147 & 146.7 & 20.96 & 148.6 & 61\% \\
    & pst-SBP & Post-Dialysis SBP & mmHg & 137.9 & 22.02 & 137.6 & 137.9 & 22.52 & 138.6 & 61\% \\
    & pre-DBP & Pre-Dialysis DBP & mmHg &78.88 & 11.9 & 78.66 & 75.62 & 11.83 & 76 & 61\% \\
    & pst-DBP &  Post-Dialysis DBP & mmHg & 77.44 & 12.05 & 77 & 73.62 & 12.53 & 73.33 & 61\% \\
    & pre-Urea & Pre-Dialysis Urea & mmol/L & 45.51 & 19.65 & 39.9 & 43.15 & 20.24 & 36.76 & 61\% \\
    & pst-Urea & Post-Dialysis Urea  & mmol/L & 15.11 & 8.526 & 12.88 & 15.02 & 8.967 & 12.79 & 67\% \\
    \hline
    & \multicolumn{2}{l}{Baseline Features}    & &  &  &  &  &  & & \\
    \hline
     & BMI & Body Mass Index & - & 21.87 & 3.633 & 21.39 & 21.83 & 4.081 & 21.60 & 57\% \\
     & Gender &  Female (0) or male (1)& - & 0.54 & 0.498 & 1 & 0.557 & 0.497 & 1 & 0\% \\
     & Age &  Age & year & 58.19 & 14.21 & 59 & 67.75 & 12.72 & 70 & 0\% \\
     & Diabetes &  Is (1) or not (0) has diabetes & - & 0.267 & 0.442 & 0 & 0.351 & 0.478 & 0 & 0\%
\\
  \hline
\end{tabular}
\end{table*}


\begin{table}[]
\centering
\caption{\textbf{Mortality Prediction Performance on Hemodialysis ESRD Dataset.} Our proposed deep-learning-based model, \mname, outperforms other SOTA baseline comparative approaches. }
    \label{tab:performance-external}
\begin{tabular}{ccccccc}
\hline
Method     & AUPRC & AUROC \\
\hline
GRU~\cite{meyer2018machine}      &  0.252(0.086)&  0.702(0.083) \\ 
Transformer~\cite{nitski2021long}   & 0.256(0.096) & 0.695(0.100) \\ 
MT-RHN~\cite{tomavsev2019clinically} & 0.275\,(0.089) & 0.735\,(0.080)\\ 
LSTM~\cite{thorsen2020dynamic}                & 0.257\,(0.085)  & 0.714\,(0.074)  \\ add static
biLSTM-FC~\cite{sung2021event}              & 0.287\,(0.082) & 0.731\,(0.078)\\ 
LR~\cite{ravizza2019predicting}   & 0.166\,(0.076) & 0.522\,(0.031)  \\ 
XGBoost~\cite{yan2020interpretable}             & 0.222\,(0.118)  & 0.518\,(0.017)  \\ 
DT~\cite{noh2020prediction}       &  0.202\,(0.034)& 0.539\,(0.023)\\ 
LightGBM~\cite{hyland2020early}            & 0.168\,(0.111) & 0.514\,(0.020) \\ 
\hline
\mname  & \textbf{0.325}$^{**}$\,(0.122)& \textbf{0.743}$^{**}$\,(0.088) \\
\hline
\end{tabular}
\end{table}

\section{Additional Interpretability Analysis}

\subsection{Additional Case Study on AI-Doctor Interaction System} 
\label{sec:system}




On the patient detail page, users can view the patient's static baseline demographic information (e.g., gender, age), dynamic trajectories of biomarkers, and prediction results.
The system automatically displays the most \textit{key} biomarkers that dominate the prediction results and provides the importance weights assigned by the model. 

\begin{figure}[]
  \centering
  \includegraphics[width=1\columnwidth]{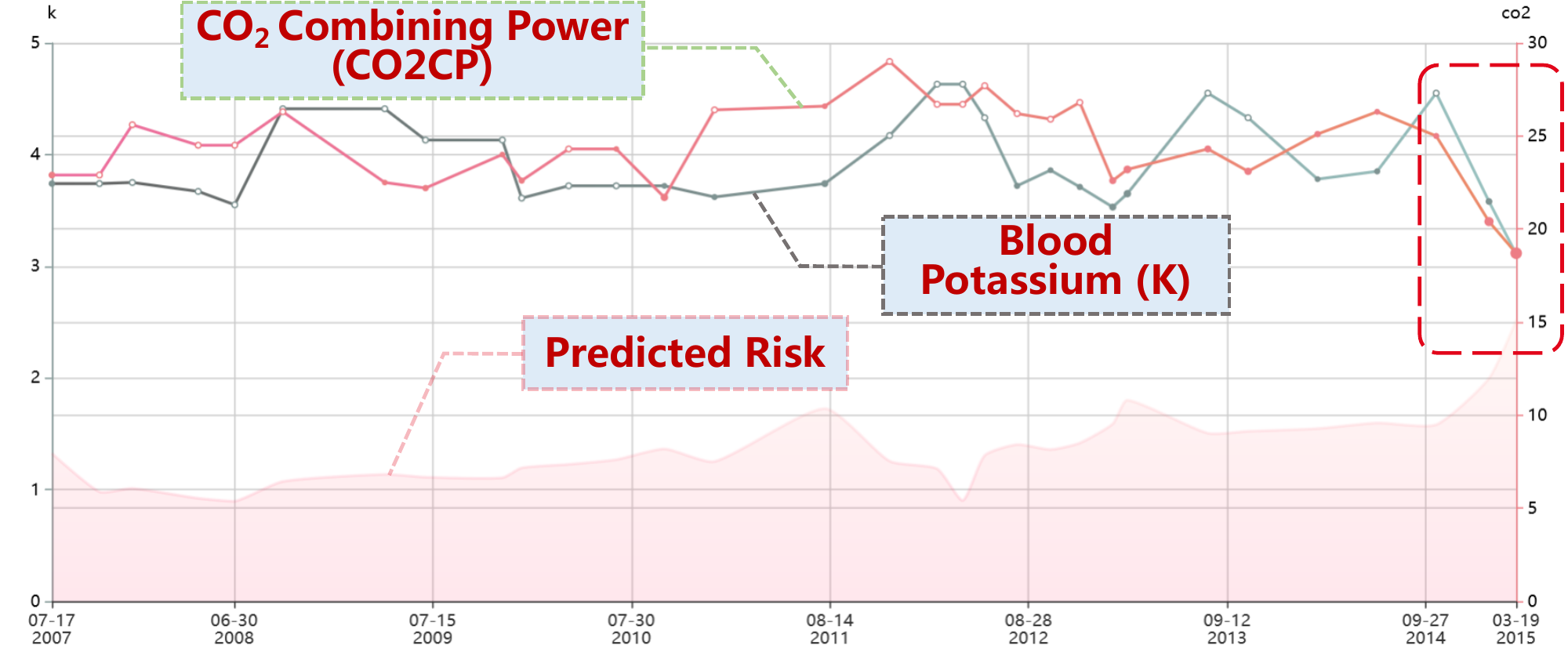}
  \caption{\textbf{Case Study III: Patient Died of Insufficient Dialysis and Sudden Heart Attack.}
  \mname pays most attention to \textcolor{pink}{CO2CP} and \textcolor{gray}{Potassium} for this patient.
 The health trajectory interactive visualization system is publicly deployed at \url{http://47.93.42.104/A1}.}
  \label{fig:215}
\end{figure}

\begin{figure}[]
  \centering
  \includegraphics[width=1\columnwidth]{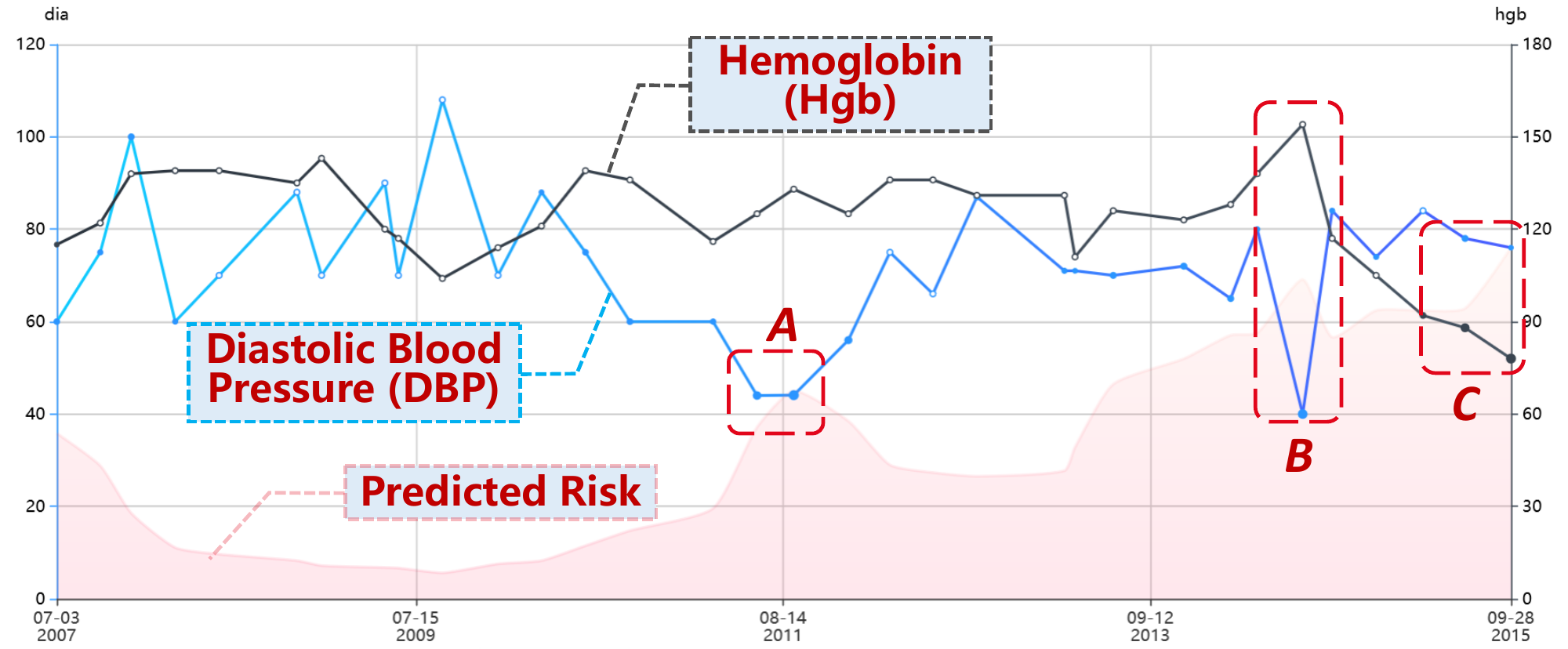}
  \caption{\textbf{Case Study IV: Patient Died of Gastrointestinal Bleeding.}
  \mname pays most attention to \textcolor{blue}{Diastolic Blood Pressure} and \textcolor{gray}{Hemoglobin} for this patient.
 The health trajectory interactive visualization system is publicly deployed at \url{http://47.93.42.104/A3}.}
  \label{fig:318}
\end{figure}

\noindent \textbf{Case III: Patient Died of Sudden Death and Insufficient Dialysis (Fig.~\ref{fig:215})}

This case was a 46-year-old female. 
She was diagnosed with diabetic nephropathy and initiated PD therapy on November 1, 2003.
Fig.~\ref{fig:215} shows the details of the previous visits and the results of the mortality risk assessment for this patient.
The model believed that the patient's risk of death rose during the last two visits, indicating a deterioration of her health. 
The model assigned high attention weights to two medical features: carbon dioxide combining power (CO2CP) and potassium (K), due to the substantial decline of these two features at the last visits. 
\mname recommended focusing on the changes of carbon dioxide combining power (CO2CP) and potassium (K), where CO2CP decreased by 6.3 mmol/L rapidly (from 25.0 mmol/L to 18.7 mmol/L, 25.2\% relative decrease), and K decreased by 1.44 mmol/L (from 4.55 mmol/L to 3.11 mmol/L, 31.6\% relative decrease).
The predicted mortality risk rapidly increased from 31.6 to 50.4 from September 2014 to March 2015.

This patient died on May 26, 2015.
The cause of death was inadequate dialysis due to abdominal leakage of dialysate and sudden heart attack. 
The model sensed changes in important physiological indicators of the patient and provided early warning of possible risk factors for the patient.
CO2CP was an indicator of acidosis, closely related to inadequate dialysis, and acidosis posed a health risk to the patient. 
The alert of potassium level could help physicians to give attention to it and treat hypokalemia by potassium supplement or hyperkalemia by necessary medical intervention.

\noindent \textbf{Case IV: Patient Died of Gastrointestinal Bleeding (Fig.~\ref{fig:318})}

This case was a 50-year-old man.
He was diagnosed with glomerulonephritis and initiated PD therapy on April 7, 2006. 
There were three risk scores peaks during the whole PD treatment period.
The first risk score peak (A) occurred on September 5, 2011.
The risk score increased from 12.1 one year before, on July 1, 2010, to 37.53 on September 5, 2011. 
During this period, AI paid attention mainly to serum albumin (Please visit the link \url{http://47.93.42.104/A3} for details.). 
On September 5, 2011, \mname found abnormal value changes in blood pressure (DBP dropped from 60 mmHg to 44 mmHg and SBP dropped from 100 mmHg to 96 mmHg) and paid much attention to DBP (26.3\%) and SBP (12\%) behind serum albumin (33.8\%).
The second risk score peak (B) was on July 24, 2014. The risk score increased to 57.59. 
The model found abnormal values of SBP (73mm Hg) and DBP (40 mm Hg). 
Attention was mainly given to SBP (28.1\%) and DBP (24.3\%).
The third risk peak (C) was on September 28, 2015. The patient risk score increased to 63.37. 
The model found an abnormal decline and low hemoglobin value (from 92g/L to 78g/L). 
The highest attention (26.1\%) was given to hemoglobin level, followed by 25.5\% of attention to serum albumin (29.1g/L). 
The patient died of gastrointestinal bleeding on November 24, 2015. 
\mname accurately captured the abnormality of blood pressure level and hemoglobin. 
Hemoglobin level was one of the main manifestations of gastrointestinal bleeding causing this patient's death.

\subsection{Additional Observation of Feature Importance Variation}
\label{sec:add_featureimportamnce}

1)-5) For analysis of albumin, diastolic blood pressure, systolic blood pressure, creatinine, and hemoglobin, please see section \ref{sec:turningpoint} in the main text.

\textbf{6) Chlorine (Cl)} Fig.~\ref{fig:feature-importance-variation}-(d).
As shown in Table~\ref{tab:feature_summary},
the importance weight of chlorine presents as a V-shaped curve with 96 mmol/L as the lowest turning point.
In the 82-96 mmol/L range, \mname pays higher attention weight to chlorine level as the chlorine gets lower and predicts a poor prognosis, marked as red dots in Fig.~\ref{fig:feature-importance-variation}-(d).
Meanwhile, in the range of 96-120 mmol/L, the model pays more attention to chlorine level as it gets higher and predicts a good prognosis.
\mname recommends raising the chloride level to above and even higher than 96 mmol/L for most PD patients.
To the best of our knowledge, the direct effect of chlorine on PD patient mortality has not been investigated by previous studies. 
Some studies find that blood chlorine level positively correlates with residual renal function \cite{li2016association}. 
The suggested level of blood chlorine by \mname is highly consistent with the traditional reference range of chlorine for outpatients, which is 96-106 mmol/L.



\textbf{7) Urea} Fig.~\ref{fig:feature-importance-variation}-(f).
The importance variation learned by \mname presents an L-shaped curve with a 20 mmol/L turning point.
The model pays more attention to urea levels in the range of 4-20 mmol/L as the urea level decreases.
When urea level drops below 10 mmol/L, the model provides more than 20\% of attention weights, and patients in this range are more likely to suffer a high mortality risk (i.e., $\hat{y}>0.5$, marked as red dots in Fig.~\ref{fig:feature-importance-variation}-(f)).
However, it nearly occupies no attention for urea levels in the range of 20-44 mmol/L.
\mname recommends maintaining urea level at 20 mmol/L or slightly higher for most PD patients. 
It will not bring much benefit if urea rises above 20 mmol/L.
This may be because the urea level indicates the nutritional status of PD patients.
The traditional recommended reference range of urea for outpatient clinics is 3.1-9 mmol/L, which is inconsistent with \mname's suggestions.
The traditional recommended range is only suitable for normal outpatients without ESRD. 
To the best of our knowledge, the recommended reference range of urea for PD patients has not been analyzed by existing works based on end-to-end deep learning methods.

\textbf{8) Calcium (Ca)} Fig.~\ref{fig:feature-importance-variation}-(g).
The importance variation of calcium presents as an L-shaped curve with a turning point of 2.5 mmol/L.
For calcium in the 1-2.5 mmol/L range, \mname pays more attention to calcium and tends to make adverse predictions as the calcium level decreases. 
For calcium over 2.5 mmol/L, the model pays nearly no attention to it. 
\mname recommends maintaining calcium level at least 2.5 mmol/L, while increasing calcium above 2.5 mmol/L may not bring much more benefits.
The traditional recommended calcium reference range for outpatient clinics is 2.25-2.75 mmol/L, which is consistent with \mname's suggestions.

\textbf{9) Sodium (Na)} Fig.~\ref{fig:feature-importance-variation}-(h).
The importance variation of sodium presents as an L-shaped curve with a turning point of 135.5 mmol/L.
For sodium level in the 121-135.5 mmol/L, \mname pays more attention to sodium and tends to make adverse predictions as the sodium level decreases. 
This is consistent with the previous study that an increased mortality rate associated with hyponatremia in PD patients \cite{al2017effect}. 
Moreover, a study in time-dependent analysis shows that, in PD patients, lower time-dependent and baseline sodium levels were independently associated with higher death risk \cite{ravel2017serum}. 
Yet, for sodium level over 135.5 mmol/L, the model pays nearly no attention to it.
Thus, \mname recommends maintaining sodium level at least 135.5 mmol/L, while increasing sodium above 135.5 mmol/L may not bring much more benefits.
The traditional recommended reference range of sodium for outpatient clinics is 135-145 mmol/L, which is highly consistent with \mname's suggestions.

\textbf{10) Potassium (K)} Fig.~\ref{fig:feature-importance-variation}-(i).
\mname pays more attention and tends to make adverse predictions as the potassium level decreases for potassium level in the range of 2.3-4 mmol/L. 
For potassium level over 4 mmol/L, the model pays nearly no attention to it.
\mname recommends maintaining potassium level at 4 least mmol/L, but further improvement may not bring benefits.
The traditional recommended reference range of potassium for outpatient clinics is 3.5-5.5 mmol/L. 
Previous studies have also reported the association between low potassium level and PD patients’ poor outcomes \cite{davies2021low,szeto2005hypokalemia}.

\textbf{11) Phosphorus (P)} Fig.~\ref{fig:feature-importance-variation}-(j).
\mname pays more attention and makes adverse predictions as the phosphate level decreases for phosphate level in the range of 0.5-1.5 mmol/L. 
AI recommends maintaining phosphate level at least 1.5 mmol/L, while further improvement may not bring benefits.
The traditional recommended reference range of phosphate for outpatient clinics is 1.1-1.3 mmol/L, which is \textbf{inconsistent} with \mname's suggestions. 
Low serum phosphorus is usually associated with poor dietary intake, and the association between low serum phosphorus and poor clinical outcome has been reported previously \cite{liu2017roles}.

\textbf{12) Carbon Dioxide Combining Power (CO2CP)} Fig.~\ref{fig:feature-importance-variation}-(k).
\mname pays more attention and makes adverse predictions as the CO2CP level decreases for CO2CP level in the range of 10.5-25 mmol/L. 
Yet, for CO2CP level over 25 mmol/L, the model pays nearly no attention to it.
Thus, \mname recommends maintaining the CO2CP level at least 25 mmol/L. 
However, further improvement on CO2CP does not help reduce mortality risk. 
The traditional recommended reference range of CO2CP for outpatient clinics is 20-29 mmol/L.
According to previous studies, low serum bicarbonate indicated acidosis in PD patients, which is associated with catabolism, malnutrition, and poor outcome \cite{szeto1998metabolic,kang1999metabolic}.

\textbf{13) Body Weight} Fig.~\ref{fig:feature-importance-variation}-(m).
\mname pays more attention and makes adverse predictions as the body weight decreases for body weight in the range of 29-59 kg. 
For body weight over 59 kg, the model pays nearly no attention to it.
\mname recommends maintaining body weight at 60 kg for most PD patients.
However, the further improvement of body weight may not bring benefits.
Malnutrition and low body weight are associated with higher mortality in peritoneal dialysis (PD) \cite{imam2021long,taylor1996adequacy}.


\textbf{14) Glucose} Fig.~\ref{fig:feature-importance-variation}-(n).
\mname pays more attention and makes good predictions as the glucose level decreases for glucose level in the range of 1-6 mmol/L. 
For glucose level over 6 mmol/L, the model pays nearly no attention to it.
Thus, AI recommends maintaining hemoglobin level not exceeding 6 mmol/L.
The traditional recommended reference range of glucose for outpatient clinics is 3.9-6.1 mmol/L, which is consistent with \mname's suggestions.


\textbf{15) Hypersensitive C-Reactive Protein (Hs-CRP)} Fig.~\ref{fig:feature-importance-variation}-(o).
Hs-CRP is a highly skewed L-shaped curve with a 16 mg/L turning point. 
For hs-CRP level 0-16 mg/L, \mname pays more attention and makes good predictions as the hs-CRP level decreases. 
For hs-CRP level over 16 mg/L, the model pays nearly no attention to it.
Thus, \mname recommends maintaining the hs-CRP level not exceeding 16 mg/L.
The traditional recommended reference range of hs-CRP for outpatient clinics is 0.5-10 mg/L, which is almost consistent with \mname's suggestions.


\textbf{16) White Blood Cell Count (WBC)}
Fig.~\ref{fig:feature-importance-variation}-(p).
WBC is considered by \mname as the most eccentric clinical feature included in the dataset.
For most patients, the importance weights assigned to WBC is below 1\%. 
The model believes that WBC is not a crucial feature for ESRD patients when conducting 1-year mortality prediction.

\section{Related Work}

\label{sec:related_work}

\begin{sidewaystable*}
\centering
  \caption{\textbf{Related Works about AI-based EMR Analysis}, especially for patients with kidney disease.
  Abbreviations in Patient Cohort: Trans (Transplant), AKI (Acute Kidney Injury), HD (Hemodialysis), CKD (Chronic Kidney Disease), IgAN (Immunoglobulin A Nephropathy), DKD (Diabetic Kidney Diseases). Abbreviations in Prediction Model: GBDT (Gradient Boosting Decision Tree), lightGBM (Light Gradient Boosting Machine), LR (Logistic Regression), ANN (Artificial Neural Network), SVM (Support Vector Machine), XGBoost (eXtreme Gradient Boosting), CAE (Convolutional AutoEncoder). Abbreviations in Interpreting Method: SHAP (SHapley Additive exPlanations). }
  \label{tab:relatedwork}
  \begin{tabular}{lccccccccc}
     \hline
Author/Year & Published  & Patient  & Prediction  & Prediction  & Information & Dynamic   & Interpret & Ante-Hoc & Adaptive \\

& Journal & Cohort & Task & Model & &  Monitor & Method & Interpret & Importance \\

\hline

Noh 2020 \cite{noh2020prediction}  & Nature Sci. R. & PD & Mortality  & Decision Tree & Static& $\times$ & Tree& $\surd$ & $\times$ \\

Zhou 2021 \cite{zhou2021prediction} & Aging Alb. NY & PD & Premature Mortality  & ANN & Static& $\times$ & Permutation& $\times$ & $\times$ \\

Chaudhuri 2021 \cite{chaudhuri2021machine} & IJMI & HD & Hospitalization & XGboost & Static& $\times$ & $\times$ & $\times$ & $\times$\\ 

Radovic 2021 \cite{radovic2022machine} & CMBBE & HD & Mortality  & SVM & Static& $\times$ & Permutation& $\times$ & $\times$ \\

Akbilgic 2019 \cite{akbilgic2019machine} & Kidney Int. R. & ESRD & Postdialysis Mortality  & Random Forest & Static& $\times$  & Tree & $\surd$ & $\times$\\

Bai 2022 \cite{bai2022machine} & Nature Sci. R. & CKD & ESRD & Naive Bayes & Static& $\times$ & $\times$ & $\times$ & $\times$\\

Makino 2019 \cite{makino2019artificial} & Nature Sci. R. & DKD & Aggravation & CAE + LR & Sequential& $\surd$ & Inverse & $\times$ & $\times$ \\

Schena 2021 \cite{schena2021development} & Kidney Int. & IgAN & ESRD & ANN & Static & $\times$ & $\times$& $\times$ & $\times$ \\

Srinivas 2017 \cite{srinivas2017big} & Ame. Jour. Trans. & Kidney Trans. & Graft Loss/Mortality  &  LR & Both & $\times$ & $\times$ & $\times$ & $\times$ \\

Liu 2021 \cite{liu2021predicting} & PLOS ONE & AKI in ICU & Mortality  & XGBoost & Static& $\times$ & Tree& $\surd$ & $\times$ \\

Kang 2020 \cite{kang2020machine} & Critical Care & CRRT for AKI & In-hospital Mortality  & XGBoost & Static& $\times$ & Tree & $\surd$ & $\times$\\

Ravizza 2019 \cite{ravizza2019predicting} & Nature Medicine & Diabetes & CKD Early Risk  & LR & Static& $\times$ & $\times$ & $\times$ & $\times$\\

\hline

Xu 2019 \cite{xu2019stratified} & MEDINFO & ICU & AKI  & GBDT& Static & $\times$ & $\times$& $\times$ & $\times$ \\

Hyland 2020 \cite{hyland2020early} & Nature Medicine & ICU & Circulatory Failure  & lightGBM& Sequential & $\surd$ & SHAP& $\times$ & $\times$ \\

Thorsen 2020 \cite{thorsen2020dynamic} & Lancet Digi. Heal. & ICU & Mortality  & LSTM& Both & $\surd$ & SHAP& $\times$ & $\times$ \\

Sung 2021 \cite{sung2021event} & JMIR Med. Info. & ICU & Mortality/AKI  & biLSTM& Both & $\surd$ & $\times$ & $\times$ & $\times$\\

Tomasev 2019 \cite{tomavsev2019clinically} & Nature & In-Patient & AKI  & Multitask RHN& Sequential & $\surd$ & $\times$ & $\times$ & $\times$\\

Yan 2020 \cite{yan2020interpretable} & Nature Mach. Int. & COVID-19 & In-hospital Mortality  & XGBoost& Static & $\times$ & Tree & $\surd$ & $\times$\\

Meyer 2018 \cite{meyer2018machine} & Lancet Res. Med. & Cardiosurgical & Complications  & GRU& Sequential & $\surd$ & $\times$ & $\times$ & $\times$\\

Raket 2020 \cite{raket2020dynamic} & Lancet Digi. Heal. & Schizophrenia & Psychosis & RNN& Sequential & $\surd$ & $\times$& $\times$& $\times$\\

Nitski 2021 \cite{nitski2021long} & Lancet Digi. Heal. & Liver Trans. & Mortality  & Transformer& Sequential & $\surd$ & Gradient &  $\times$ & $\times$ \\

Rank 2020 \cite{rank2020deep} & NPJ Digital Med. & Cardiosurgical & AKI & RNN& Sequential & $\surd$ & $\times$& $\times$ & $\times$ \\

     \hline 
\textbf{Ours} 2022 & - & PD (HD) & Mortality & \mname & \textbf{Both} & \pmb{$\surd$} & \mname & \pmb{$\surd$} & \pmb{$\surd$}\\
     \hline
\end{tabular}
\end{sidewaystable*}

Over the past ten years, there has been a massive explosion in the amount of digital information stored in electronic medical records, which opens a door for researchers to make secondary use of these records for various clinical applications.
At the same time, with the development of artificial intelligence,
machine learning and deep learning-based models have shown the capability to perform renal-disease-related clinical predictions, including acute kidney injury risk prediction~\cite{tomavsev2019clinically,sung2021event,xu2019stratified,rank2020deep}, graft loss prediction~\cite{srinivas2017big} and mortality prediction~\cite{sung2021event,srinivas2017big,noh2020prediction,akbilgic2019machine,liu2021predicting,zhou2021prediction,radovic2022machine,kang2020machine}.
For instance,
Bai et al.~\cite{bai2022machine} proposed a study to assess the feasibility of machine learning (ML) in predicting the risk of end-stage kidney disease (ESKD) for patients with CKD.
~\cite{schena2021development} developed a deep-learning-based prediction model for end-stage kidney disease (ESKD) in patients with primary immunoglobulin A nephropathy (IgAN). 
Noh et al.~\cite{noh2020prediction} conducted the 5-year mortality risk prediction task for peritoneal dialysis patients using the decision tree model.
Xu et al.~\cite{xu2019stratified}, Ravizza et al.~\cite{ravizza2019predicting} and Chaudhuri et al.~\cite{chaudhuri2021machine} utilized patient static information to conduct the prediction for the progression of renal diseases, such as AKI, CKD and hospitalization.
Besides, Akbilgic et al.~\cite{akbilgic2019machine}, Liu et al.~\cite{liu2021predicting}, Zhou et al.~\cite{zhou2021prediction}, Radovic et al.~\cite{radovic2022machine} and Kang et al.~\cite{kang2020machine} used different ML-based methods to predict the mortality risk of patients with kidney-related diseases.

However, there are still some critical issues that have not yet been thoroughly addressed by existing work in terms of the following three issues, as shown in Table~\ref{tab:relatedwork}.

\textbf{ $I_1$: Perform dynamic mortality prediction at each follow-up visit based on the effective utilization of both sequential medical records and the baseline demographic information.}
Most above-mentioned existing renal disease-related works only utilize static records.
Such models cannot learn non-linear progression patterns from high-dimensional longitudinal EMR datasets, capturing the health status variation trajectory, limiting their prediction performance and applications in quality improvement initiatives or data-driven clinical decision-making processes. 
It requires significant efforts from clinicians, healthcare institutions and model developers to collect data and build longitudinal models for such long-term predictive tasks.

Some works attempted to model the dynamic information of patients.
For example,
Makino et al.~\cite{makino2019artificial} constructed a predictive model for diabetic kidney diseases (DKD) using AI, processing natural language and longitudinal data of diabetes patients. 
Rank et al.~\cite{rank2020deep} developed a deep-learning-based real-time algorithm to predict postoperative AKI prior to the onset of symptoms and complications.
Tomavsev et al.~\cite{tomavsev2019clinically} and Hyland et al.~\cite{hyland2020early} proposed DL-based models to conduct dynamic monitoring for the risk of AKI and circulatory failure on in-hospital patients with kidney disease.
These work dynamically predict patients' risk through different time-series models based on deep learning and realize the dynamic monitoring of the health status of patients with kidney disease. 
However, the baseline data of patients with renal disease is also essential in diagnosing and treating. 
These researches have limitations in combining the static baseline information and dynamic data of patients, thence have difficulties comprehensively integrating the patient information for evaluation.

Thence, Srinivas et al.~\cite{srinivas2017big} combined the static and dynamic data of patients who received a kidney transplant and conducted the graft loss and mortality prediction via LR.
Sung et al.~\cite{sung2021event} have used biLSTM-FC to perform the clinical event prediction (death, sepsis, and acute kidney injury), where biLSTM and fully connected layers are employed to embed the dynamic features and static features correspondingly.
However, these works briefly concatenate the static and sequential information by the hidden units.
The static information does not guide the individual health status representation learning or help the model adaptively assign weights to the features.


\textbf{$I_2$: Provide fine-grained interpretability for each patient individually by selecting key features which contribute the most to the mortality prediction (patient-level interpretability) and achieve high prediction performance simultaneously.}
Although deep learning has achieved huge success in many domains, lack-of-explainability remains one serious drawback for the neural network.
An interpretable model is essential for clinical decision support applications as the predictive results need to be understood by clinicians to adopt individualized treatment and extract medical knowledge.
However, the model interpretability has not been fully discussed in most renal disease-related works.
The decision-making process in such deep models is a black box and fails to provide human-understandable interpretability.

Several researchers have explored the interpretability in the medical feature via tree-based strategy.
For example, Noh et al. \cite{noh2020prediction} assessed mortality risk prediction in PD patients using decision tree algorithms.
Akbilgic et al. \cite{akbilgic2019machine} implemented a random forest method to predict outcomes of ESRD patients after dialysis initiation.
However, the prediction performance of these static information-based methods is limited due to the deficiency of effective advanced clinical feature extraction.

Some recent works apply the SHapley Additive exPlanations (SHAP) \cite{hyland2020early,thorsen2020dynamic}, feature permutation \cite{zhou2021prediction,radovic2022machine}, and inverse analysis \cite{makino2019artificial} strategies to provide the post-hoc interpretability.
For example, Makino et al. generated the time-series data pattern by inverse analysis.
However, these interpreting methods usually only provide coarse-grained analysis and may face a difficult trade-off between the network complexity and prediction performance.
As a result, it is still challenging to provide satisfying interpretability and achieve high prediction performance simultaneously. 

\textbf{$I_3$: Adaptively analyze the importance of each feature along with the variation of its value (feature-level interpretability) to provide medical advice and extract knowledge.}
The interpretability shown in most of the existing EMR analysis works mainly focuses on tree-based interpretability and some forms of post-hoc interpretability.
The tree-based analysis can only provide a fixed decision process for all patients and face a deficiency in sequential information utilization.
To the best of our knowledge, none of the existing PD/HD/ESRD-related works explicitly provide the adaptive feature importance, analyze the changes of feature importance with its values, and extract medical advice based on ante-hoc interpretability in a deep end-to-end model.

\section{Data Availability}

We develop a practical AI-Doctor interaction system to visualize the trajectory of patients' health status and risk indicators. Our developed health trajectory visualization system with anonymous case studies (patient ID from A1 to A20) is publicly available at \url{http://47.93.42.104/A8}. Visualization of the importance of the features is available at \url{http://47.93.42.104/statistics/feature}. More data that support the findings of this study are available from the corresponding author upon reasonable request.

We release our code at \url{https://github.com/Accountable-Machine-Intelligence/AICare}.
Users can upload the data online to get the prediction results immediately \url{http://47.93.42.104/predict} or download the code to train the model based on their dataset offline.

\section{Competing Interests}

The Authors declare no Competing Financial or Non-Financial Interests.

\section{Author Contributions}

Liantao Ma and Chaohe Zhang proposed the main architecture of the AICare model. The research process was designed by Liantao Ma, Junyi Gao, and Yasha Wang. Chaohe Zhang and Xianfeng Jiao conducted the experiment. Liantao Ma and Zhihao Yu visualized the results. The AI-Doctor interaction system was designed by Liantao Ma, and implemented by Xianfeng Jiao and Zhihao Yu. The data was collected by Tao Wang, Wen Tang and Xinju Zhao. Junyi Gao and Liantao Ma conducted the data pre-processing. The prediction task is formulated by Liantao Ma, Junyi Gao, Wen Tang and Yasha Wang. Wen Tang and Xinju Zhao extracted the medical findings.

\section*{Acknowledgment}

This work was supported by the National Natural Science Foundation of China (No.62172011), PKU-Baidu Foundation (2020BD030), China International Medical Foundation (Z-2017-24-2037), and China Postdoctoral Science Foundation (2021TQ0011).
This work was partly supported by the UK EPSRC project on Offshore Robotics for Certification of Assets (ORCA) [EP/R026173/1] and the Australian Research Council under Grant DE200101439.
Junyi Gao acknowledges the receipt of studentship awards from the Health Data Research UK-The Alan Turing Institute Wellcome PhD Programme in Health Data Science (Grant Ref: 218529/Z/19/Z).

\ifCLASSOPTIONcaptionsoff
  \newpage
\fi

\bibliographystyle{IEEEtran}
\bibliography{bare_jrnl_compsoc}

\end{document}